\documentclass[preprint,12pt]{elsarticle}

\let\today\relax
\makeatletter
\def\ps@pprintTitle{%
    \let\@oddhead\@empty
    \let\@evenhead\@empty
    \def\@oddfoot{\footnotesize\itshape
         {} \hfill\today}%
    \let\@evenfoot\@oddfoot
    }
\makeatother

\usepackage{caption}
\usepackage{subcaption}
\usepackage{booktabs}
\usepackage{hyperref}
\usepackage{natbib}
\usepackage{amssymb}
\usepackage{amsmath}
\usepackage{lineno}
\usepackage{url}
\usepackage{xcolor}
\usepackage{afterpage}

\journal{Artificial Intelligence}

\begin{document}

\begin{frontmatter}



\title{Counterfactual Explanations for Misclassified Images: \\How Human and Machine Explanations Differ}


\author[label1,label2,label3]{Eoin Delaney\corref{cor1}}
\author[label1,label3]{Arjun Pakrashi}
\author[label1,label2,label3]{Derek Greene}
\author[label1,label2,label3]{Mark T. Keane}

\cortext[cor1]{Corresponding author - \textbf{eoin.delaney@insight-centre.org}}

\affiliation[label1]{organization={School of Computer Science, University College Dublin},
            addressline={Belfield},
            city={Dublin},
            country={Ireland}}

\affiliation[label2]{organization={Insight Centre for Data Analytics},
            addressline={Belfield},
            city={Dublin},
            country={Ireland}}

\affiliation[label3]{organization={VistaMilk SFI Research Centre},
            addressline={Belfield},
            city={Dublin},
            country={Ireland}}


\begin{abstract}

Counterfactual explanations have emerged as a popular solution for the eXplainable AI (XAI) problem of elucidating the predictions of black-box deep-learning systems due to their psychological validity, flexibility across problem domains and proposed legal compliance. While over 100 counterfactual methods exist, claiming to generate plausible explanations akin to those preferred by people, few have actually been tested  on users ($\sim$7\%). So, the psychological validity of these counterfactual algorithms for effective XAI for image data is not established. This issue is addressed here using a novel methodology that (i) gathers ground truth human-generated counterfactual explanations for misclassified images, in two user studies and, then, (ii) compares these human-generated ground-truth explanations to computationally-generated explanations for the same misclassifications. Results indicate that humans do not “minimally edit” images when generating counterfactual explanations. Instead, they make larger, ``meaningful" edits that better approximate prototypes in the counterfactual class.

\end{abstract}



\begin{keyword}
XAI \sep Counterfactual Explanation \sep User Testing



\end{keyword}

\end{frontmatter}



\section{Introduction} 
\label{}
As Artificial Intelligence (AI) is increasingly used in everyday life for high-stakes decision-making, many new roles have emerged for eXplainable AI (XAI) \cite{adadi2018peeking, doshi2017towards, goodman2017european, sokol2019CFsafeAI}.
For instance, in computer vision systems, explanations can help to debug black-box models (e.g., showing why images were misclassified) \cite{ross2018improving,bauerle2018training}, to audit system safety (e.g., why a self-driving car misidentified a postbox as a red light \cite{goyal2019counterfactual}), to assess fairness and bias (e.g., why one person’s face was cropped from an image over another’s \cite{birhane2022auditing}) and, even, to provide novel domain insights (e.g., identifying mass legions in digital mammography \cite{barnett2021case}).

In computer vision systems, many diverse strategies have been advanced to explain model predictions \cite{lipton2016mythos,adadi2018peeking,guidotti2018survey} using, for instance, saliency maps \cite{zhou2016learning, selvaraju2017grad}, feature importance \cite{ribeiro_why_2016,lundberg2017unified}, prototypes \cite{kim2016criticism,Rudin2019}, and factual \cite{sormo_explanation_2005,keane2019case}, counterfactual \cite{miller2019explanation,Byrne2019} or semifactual examples \cite{kenny2021explaining}.
Saliency maps have been extensively used in image classification to highlight ``important regions" in the input image by using back-propagation and up-sampling to generate an activation map \cite{zhou2016learning, selvaraju2017grad}. In a similar fashion, explanations using model-agnostic feature importance (e.g., LIME \cite{ribeiro_why_2016} and SHAP \cite{lundberg2017unified}) aim to show the features or super-pixels of an instance that contributed most to a prediction. Indeed, recently, to achieve better fidelity to the underlying black-box model \cite{adebayo2018sanity}, there has been a move away from input-level features (pixels or segments) to more high-level concepts (e.g., stripes in an image of a zebra) based on extracted latent features (e.g., as concept activation vectors \cite{kim2018interpretability,ghorbani2019towards, chen2020concept, zhang2021invertible}).  By the same token, example-based explanations try to leverage meaningful, concept-level units (e.g., as prototypes, factual, counterfactual or semifactual examples).  Here, we focus on the psychological validity of counterfactual methods, although our results are also shown to have significant implications for prototype techniques.

Counterfactual explanations have received significant attention in the XAI literature (for reviews see \cite{guidotti2018survey,verma2020counterfactual,karimi2020algorithmic,keane2021if}) as they provide ``what if" explanations that use a contrasting case to show how a prediction would change \textit{if} the input features had been different \cite{goyal2019counterfactual,Guidotti2019,miller2019explanation,karimi2020algorithmic,keane2021if}. For instance, when a two-year loan application for \$5k is refused by an automated system and the user asks ``Why?", the counterfactual explanation might suggest ``If you had asked for a \$4.5k loan over a one-year term, then you would have been granted the loan". In this case it can be seen that changing the \textit{loan-amount} and \textit{loan-duration} features can flip the decision to the user's desired outcome (i.e., \textit{loan-granted}).  In the image domain, a corresponding counterfactual could be used to audit a misclassification by a black-box deep learner; for instance, when a self-driving car misidentifies a postbox as a red light and the user asks ``Why?", the counterfactual explanation might suggest "if the round-red-postbox had been a square-red-postbox, it would \textit{not} have been misidentified as a traffic light".   More formally, given a black-box classifier $b$ and $I$ as some to-be-explained query image with the predicted class $b(I) = y$, then $I'$ is a candidate counterfactual explanation when $b(I') = y'$, where $y$ and $y'$ are contrasting classes (see e.g., \cite{goyal2019counterfactual}).

The current AI interest in counterfactual methods has been boosted by philosophical proposals about their centrality in causality \cite{lewis2013counterfactuals,woodward2005making}, psychological findings that they are important to people's understanding of causes \cite{miller2019explanation,Byrne2019,byrne2007rational,mueller2019explanation,lagnado2013causal} and legal analyses suggesting they are GDPR compliant \cite{wachter2017counterfactual}.  Indeed, there are now 100+ counterfactual methods in the XAI literature, all of which claim to generate the \textit{plausible} counterfactual explanations people need to understand AI systems \cite{keane2021if}. However, most of these plausibility claims are based on intuition rather than hard psychological evidence \cite{barocas2020hidden,leavitt2020towards}.  As with much of the XAI literature \cite{keane2019case, anjomshoae2019explainable}, the user testing of methods remains uncommon; Keane et al.~\cite{keane2021if} found that only $\sim$7\% of counterfactual methods specifically user-tested proposed functionalities. So, we really do not know which, if any, of these counterfactual methods \textit{really} generates explanations that people find plausible.

\textit{}Accordingly, in this paper, we advance a novel methodology to address this significant gap in the counterfactual XAI literature.  Stated simply, we ask people to counterfactually explain images misclassified by an AI system and then we compare these explanations to those produced by several benchmark counterfactual methods for the same misclassifications. These ground-truth human explanations allow us to assess definitively whether the machine-generated counterfactuals are, indeed,  plausible. To presage our results, the evidence shows that they are \textit{not}, that people actually prefer counterfactual explanations that rely more on prototypes from a contrasting class, rather than minimally perturbed instances close to decision boundaries.


\subsection{Contributions \& Outline of Paper}
This paper aims to make significant progress in providing novel results and a better psychological grounding for the use of counterfactual explanations in XAI.  As such, we make several novel contributions to the field:
\begin{itemize}
\item We provide a critical analysis of the notion of plausibility in the counterfactual XAI literature, showing how different intuitive claims for it can be mapped to evaluative metrics measuring proximity, representativeness, and prototypicality.

\item We provide an up-to-date survey of the main user-study findings on counterfactual explanations in XAI including a critical analysis that reveals the system-centered nature of this work.

\item We advance a novel user-centered methodology for collecting counterfactual explanations used by people and show how this ground truth can be related to matched outputs from computational counterfactual methods.

\item We report new results on the specific divergences that occur between human and machine explanations when evaluation metrics for proximity, representativeness, and prototypicality are applied; detailing the major implications they have for current methods in the field.
\end{itemize}

\noindent In the next section, we review the related work on counterfactual methods in XAI, their plausibility claims, and the evaluation metrics used to test these claims (see section 2).  Then, we survey  the main findings emerging from the scarce user tests that have been performed on the efficacy of counterfactual explanations methods, introducing the proposal that they are predominantly system-centered (see section 3).  We then sketch our novel methodology for collecting a ground truth of human counterfactual explanations for misclassifications in two benchmark image datasets (MNIST \cite{lecun1998mnist} and QuickDraw Doodles \cite{cai2019effects}; see sections 4).   Section 5 presents our comparative study of counterfactual explanation, detailing the correspondences between human-generate and machine-generated counterfactuals. Finally, we discuss the implications of these results for counterfactual explanations, in particular and, more generally, for he explanation of AI systems.

\section{Related Work I: Plausibility of Counterfactual Explanations}
\label{Related Work I}

\noindent In the current section, we do not attempt to survey the 100+ counterfactual methods that exist in the XAI literature (see \cite{karimi2020algorithmiccf,verma2020counterfactual,keane2021if} for good surveys). Rather, we focus on the literature of most relevance to the current study to contextualise profile, and detail the methods tested here.  Most counterfactual methods claim, typically on intuitive grounds, to produce \textit{plausible} counterfactuals; that is, counterfactuals that people would find realistic, comprehensible and appropriate for explanatory purposes. However, different methods differ in what they see as the basis for this plausibility, proposing that it relies on some deeper aspect of the counterfactual-computations made by the method.  For instance, many counterfactual methods, since Wachter et al.'s seminal paper \cite{wachter2017counterfactual}, argue that \textit{proximity} is the key to plausibility; that counterfactuals that are close to the query and to the decision boundary of the counterfactual class lead to the plausible counterfactuals that people require. Focusing on proximity also induces the generation of sparse counterfactuals that modify few aspects of the original query instance, thus making them easier for people to comprehend \cite{keane2020good}.  Other methods have proposed that plausibility owes more to \textit{representativeness}, that the counterfactual explanation needs to be representative of the domain to be plausible (e.g., \cite{laugel2019dangers}). Still others have argued that plausibility hinges on using semantically-meaningful features that represent the central tendency of instances in the class, specifically, those captured by \textit{prototypes} (e.g., \cite{VanLooveren2019}). In this section, we introduce and critically review the methods tested in the current study, from these different plausibility perspectives (i.e., proximity, representativeness, prototypicality) in terms of (i) the perspective on plausibility taken, (ii) the details of the methods tested, (iii) the evaluation metrics used to assess this type of plausibility.

\label{sec:proximity}
\subsection{Proximity: Plausibility as Counterfactual Closeness}
David Lewis' influential philosophical analysis \cite{lewis2013counterfactuals}  proposed that counterfactuals captured the closest possible world in which some target event minimally differed. In the last few years, this view that plausible counterfactual explanations depend on balancing the distance from the query and decision boundary has, perhaps, been the dominant paradigm in XAI. This idea underlies so-called \textit{proximity methods} that aim to find the minimal changes or Min-Edits to a query-instance's features to generate one with a contrasting predictive outcome \cite{miller2019explanation,wachter2017counterfactual, Mothilal2020, keane2020good, VanLooveren2019}.

The earliest methods in this literature relied on proximity to find Nearest Unlike Neighbours (NUNs) in the dataset as counterfactual explanations \cite{nugent2005case,martens2014explaining}, though this solution makes the explanation process overly dependent on the availability of suitably-close instances in the dataset. Wachter et al.'s \cite{wachter2017counterfactual} seminal contribution, the so-called \textit{Min-Edit} method, introduced the major innovation of generating synthetic counterfactual instances to provide much better explanatory coverage, though generating such instances does lead to other issues (e.g., invalid data-points being proposed as explanations).   In passing it should be said that in Psychology identifying plausibility with similarity has not been commonly proposed (for a critique see \cite{keane2021if}, though see also \cite{costello2000efficient}).

\subsubsection{Testing Proximity: Using Min-Edit}
\noindent  The Min-Edit method searches a space of perturbations of the query-instance using gradient descent, applying a loss function that balances the closeness of the counterfactual to the query against making minimal feature-changes required to deliver a prediction change \cite{wachter2017counterfactual}. So, this method aims to generate a counterfactual explanation by minimizing:
\begin{equation}
    (b_{t}(I')-p_{t})^2 + \lambda \| I - I' \|_{1}
\end{equation}
The first loss term pushes the predicted class probability of the candidate counterfactual $b_{t}(I')$ towards a target $p_{t}$, while the second term minimizes the Manhattan distance between the query and counterfactual to promote proximate and sparse solutions. The Lagrangian multiplier, $\lambda$, acts as a balancing term.

This method has been hugely influential in the recent counterfactual literature and is the recognised baseline for many algorithmic comparisons.  Furthermore, many subsequent counterfactual methods have extended this approach with additional constraints to improve sparsity \cite{dandl2020multi,grath2018interpretable}, diversity \cite{mothilal2020explaining}, and to include causal models \cite{karimi2020algorithmic}.  Other methods have tried to tackle the invalid-data-point issue by including representativeness ideas, using auto-encoders or generative models to ensure that the counterfactual lies close to the data manifold \cite{joshi2019towards, VanLooveren2019, kenny2020generating, Singla2020Explanation, dhurandhar2018explanations}.

In the present comparative study, we use Min-Edit as a baseline for this approach to counterfactual XAI and implement it using \cite{alibi}. The standard evaluation metrics for assessing plausibility-as-proximity are the L1 and L2 norms, applied to the query-explanation pairings produced by a given explanation method, where lower distances are taken to indicate the success of the method.

\subsection{Representativeness: Plausible Explanations Are Within-Distribution}
\label{sec:within}
\noindent Part of the motivation behind early counterfactual methods using Nearest Unlike Neighbours (NUNs) was the recognition that explanations would be plausible if they were representative and within the domain.  By definition, these methods cannot generate an out-of-domain counterfactual explanation, as NUNs are typically within the training distribution.  However, as we said earlier, if queries have few or no close NUNs, this can still lead to low-quality or failed explanations \cite{keane2020good, smyth2022few}. Recent perturbation methods (such as Min-Edit) deal with this coverage issue by generating synthetic counterfactuals. However, these methods can risk generating instances that are either unrealistic, or not perceptibly different from the query instance \cite{ keane2021if, joshi2019towards, dhurandhar2018explanations, russell2019efficient, delaney2021uncertainty}.  Indeed, Laugel et al. \cite{laugel2019dangers} showed that for some domains these methods could result in 30\% of explanations being out-of-distribution.  Such concerns led to a re-emphasis on representativeness, to ensure that plausible explanations are within-domain and, if possible, not out-of-distribution.  Indeed, this representativeness perspective on plausibility is much closer to proposals made in psychology, where it is often cast as coherence or consistency with prior knowledge in a domain \cite{forster2020evaluating, connell2004plausibly, connell2006model, costello2000efficient, klein2021plausibility}.

 \subsubsection{Testing Representativeness: CEM-PN, VLK \& Revise}

\noindent  Many counterfactual techniques have supplemented proximity-based methods to take representativeness into account, often by using auto-encoders or generative models. The goal is to ensure that plausible counterfactuals that belong to the data distribution are preferred \cite{kenny2020generating, dhurandhar2018explanations, joshi2019towards}. In the current study, we selected three methods taking this approach, based on their recognition in the area, their ability to handle image data, and open-source availability of their code (see \cite{dhurandhar2018explanations,VanLooveren2019,joshi2019towards}).  Here we briefly describe each in turn.

\textbf{CEM-PN} \cite{dhurandhar2018explanations} computes pertinent negatives using an objective function that contains an elastic net $(\beta L_{1} + L{2})$ regulariser to select features to alter via perturbation whilst keeping the perturbations sparse. An autoencoder is leveraged to ensure that the generated explanations lie close to the data manifold through minimizing the $L_{2}$ reconstruction error. This method was implemented using \citep{alibi}.

\textbf{VLK} \cite{VanLooveren2019} generates a counterfactual by minimising a multi-objective loss function defined by
\begin{equation}
    Loss = c L_{pred} + \beta L_{1} + L_{2} + L_{AE} + L_{proto}
\end{equation}
where the first term encourages the perturbed instance to belong to the counterfactual class. The elastic net regulariser $\beta L_{1} + L_{2}$ aims to ensure sparsity and proximity in the generated instance. $L_{AE}$ is the reconstruction error of the candidate counterfactual instance, which is minimized to encourage the counterfactual to belong to the training data distribution. To guide the counterfactual-instance towards the distribution of the perturbed class, the $L_{2}$ distance between it and the counterfactual class prototype is minimised in the $L_{proto}$ term. Following \cite{VanLooveren2019}, the encoder from $L_{AE}$ is used to retrieve class prototypes. Specifically, the counterfactual class prototype is defined as the average encoding of the five nearest instances (according to Euclidean distance) in the latent space with the same counterfactual class label.

\textbf{Revise} \cite{joshi2019towards} relies on a generative model that is a decoder of a variational autoencoder (VAE) trained on the training data. The overall idea is to minimise the function
\begin{equation}
    \ell(b(G(\textbf{z'})),t) + \lambda \| G(\textbf{z'}) - \textbf{I} \|_{1}
\end{equation}
where $b$ is the classifier, $t$ is the target, $\ell$ is some loss function, and $G$ is the generative model. To find a $z'$ that minimises the loss, $z$ is initialised to the encoding of the original input $I$. Then the gradient of the loss in the latent space is computed and the algorithm iteratively takes small steps in that space until the prediction changes to the target. Since the resulting counterfactual $I' = G(z')$ is produced by the generative model, it can be \textit{dissimilar} to $I$. Although the L1 norm is known to encourage sparsity, the algorithm may not necessarily generate sparse solutions, as the changes occur in the latent space rather than the input space. The method may also fail to generate explanations in some cases. For instance, \cite{holtgen2021deduce} reported a 22\% failure rate on MNIST images. This method is implemented using code from \cite{holtgen2021deduce}, with the recommended hyperparameters of $\lambda$ = 1 and gradient step $\delta$ = $10^{-5}$. As in \cite{joshi2019towards, holtgen2021deduce}, we use cross entropy loss for $\ell$.

In the present study, these three methods are used as benchmark approaches for this representativeness approach to counterfactual XAI, with all being implemented using \cite{alibi}. Several different evaluation metrics, typically out-of-distribution measures, have been used to assess plausibility-as-representativeness, applied to the counterfactual-instances produced by a given explanation method. We use MC-Dropout, IM1, 10-LOF and R\%Sub to perform these evaluations (see Section \ref{Representativeness Metrics}).

\subsection{Testing Prototypicality: MMD-Critic \& Grad-Cos}
\label{sec:closeness}
A final plausibility claim on counterfactual explanations is the idea that they should be within-class representative, which can be reflected in prototypicality measures; another consideration that arises from issues with perturbation methods in image domains. As proximity methods try to find counterfactuals which lie just over the decision boundary (minimising distract from the query), this often means that they are not particularly representative of the counterfactual class, or indeed the original query-class either.  Indeed, as has been discovered in adversarial learning \cite{goodfellow2015_adversarial}, small pixel-changes to a query can sometimes result in a change of class, even when these pixel-changes are actually imperceptible to humans \cite{kenny2021explaining, goodfellow2015_adversarial} (see Figure \ref{fig:mnist} for examples). Clearly, such counterfactuals are not plausible explanations for people.

One solution to this problem is to generate the  counterfactuals in a latent space using higher-level features (e.g., conceptual \cite{goyal2019counterfactual} or exceptional features \cite{kenny2021explaining}). Indeed, some have gone further, arguing that it is better to instead leverage instances that are maximally representative of a class (i.e., \textit{class prototypes}) \cite{kim2016criticism, chen_this_2018, kim2014bayesian, li2017prototypescbr, barnett2021case}, or to interpret a model's prediction using human-friendly concepts \cite{kim2018interpretability,ghorbani2019towards, chen2020concept, zhang2021invertible}).

Accordingly, we also measure the similarity of generated counterfactuals to computed prototypes from the counterfactual class (using Grad-Cos \cite{charpiat2019input} on prototypes from MMD-Critic \cite{kim2016criticism}). MMD-critic \cite{kim2016criticism} is implemented to compute prototypes by minimizing the maximum mean discrepancy between the prototype distribution and the data distribution using a kernel density function (full technical details can be found in Appendix B). This evaluation aims to determine whether generated counterfactuals are actually close to prototypes for the counterfactual class.  Recall, Min-Edit methods try to find instances that are close to the query but just over the decision boundary in a contrasting class; hence, they are much less likely to be representative members of this counterfactual class. To evaluate prototypicality we used  the Grad-Cos metric to determine whether machine-generated and human-generated counterfactuals are close to prototypes (for details see Section \ref{prototypicality_metric}).

\section{Related Work II: User Tests of Counterfactual XAI}
\noindent The recent rapid algorithmic advances in counterfactual XAI have not been paralleled by a significant program of user testing. One survey of user studies reports that only 31\% (36 out of 127 papers) of counterfactual papers conducted any form of user evaluation and only 7\% competitively test alternative algorithms \cite{keane2021if}.  Furthermore, many of these studies are vitiated by poor/non-reproducible experimental designs, imbalanced material sets, low/inappropriate numbers of participants, and inappropriate/absent statistical analyses. So, very few studies report solid evidence on the efficacy of counterfactual explanations in XAI.  Indeed, as we shall see, most of these user studies adopt a very system-centered approach, in which users are cast as passive recipients of machine-generated explanations.  In the following subsections, we critically assess the main findings from this literature under the headings of findings on (i) the general efficacy of counterfactual explanations, (ii) specific comparative tests of methods, (iii) tests of image-based methods.

 \subsection{Counterfactual Explanations: General Efficacy}
\noindent Most user studies on counterfactual XAI focus on tabular datasets and, simply, attempt to establish in broad terms whether or not explanations impact user behaviour \cite{dodge2019explaining,lim2009and,Warren2022FeaturesXAI_IJCAI,lucic2020does}.  Methodologically, they test whether the provision of a counterfactual explanation to an algorithmic decision has some/any effect on user behaviour by comparison to no-explanation controls and/or some other explanatory method (e.g., causal explanations or case-based explanations \cite{keane2019case,Warren2022FeaturesXAI_IJCAI}).

For instance, an early study by Lim et al. \cite{lim2009and}  gave separate groups of people different explanation-types -- What-if, Why-Not, How-to and Why explanations -- and found that that all of the explanation-interventions improved performance relative to no-explanation controls. However, the Why-Not counterfactual explanations did no better than the other explanation-types.   In a similar vein, Dodge et al. \cite{dodge2019explaining} assessed four different explanation strategies (e.g., case based, counterfactual, and two forms of global explanation) for biased/unbiased classifiers and found that counterfactual explanations had the greatest impact on user responding (although these tests used quite a small set of materials). Similarly, using a simulated drink-driving-advice app, Warren et al. \cite{Warren2022FeaturesXAI_IJCAI} found that counterfactual explanations improved people's understanding of the app's predictions, over causal explanations and no-explanation controls; they had participants predict outcomes for unseen instances in a testing phase after a training phase in which they were shown the app's decisions.

However, other studies have found less evidence for the efficacy of counterfactual explanations.  In a carefully-controlled study on a simulated app to advise diabetics about insulin treatments, van der Waa et al. \cite{vanderwaa2021evaluating} found that counterfactual and example-based explanations did not improve people's knowledge of the domain over no-explanation controls (as measured by users' predictive accuracy on test cases). Worryingly, this study also found that people followed the app's advice even when it was incorrect.   The latter findings raises the worrying prospect that people can be increasingly satisfied with a system when it is explained to them, without gaining any insight into how the system works (which has been raised as an issue for XAI \cite{buccinca2020proxy,Warren2022FeaturesXAI_IJCAI}). Other work \cite{lage2019evaluation} that asked people to predict/simulate what a model might do under different inputs found that tasks using counterfactuals elicited longer response times, were rated as being more difficult, and showed reduced accuracy in users. Lucic et al. \cite{lucic2020does} found that people were less accurate on tasks that asked them to introduce a counterfactual change for an instance than when they were asked to predict outcomes from the input features of the instance.  These latter studies point to an added cognitive overhead in processing counterfactuals that has been long-noted in psychology \cite{byrne2007rational,Byrne2019}.

So, overall, this literature shows mixed support for the efficacy of counterfactual explanations.  Of course, part of the problem here may arise from paucity of good studies in the literature. In time, as more studies are done, we may be able to reach a better understanding of what it is that makes counterfactual explanations work in some contexts and not in others. However, it could also be the case that people are not being
properly tested in these studies, that the orientation taken is too much from the machine-perspective than that of  users.

\subsection{Comparative Tests of Counterfactual Methods}
\noindent Apart from user tests on the general efficacy of counterfactual XAI,  given the differnet proposals on the plausibility of different algorithms, we also require comparative tests between methods.  However, even fewer user studies have done such tests \cite{akula2020cocox,forster2020evaluating,forster2021capturing,kirfel2021whatifandhow,kuhl2022keep}.  Forster and colleagues \cite{forster2020evaluating,forster2021capturing} have performed a series of unique studies that asked participants to assess counterfactuals produced by different algorithms or parametric-variations of a given algorithm \cite{forster2020evaluating,forster2021capturing}. They compared a gradient-free optimisation method against a popular gradient-based method \cite{wachter2017counterfactual} asking people to rate the typicality, realism and suitability of generated counterfactual explanations (i.e., proxy metrics for representativeness). The results showed that their optimisation method, did better on all metrics than the popular Wachter et al. \cite{wachter2017counterfactual} perturbation method. Kuhl et al. \cite{kuhl2022keep} also compared methods that differed in how they selected counterfactual instances (i.e., perturbation methods with and without density computations) but found no evidence that people could distiguish them.  Akula et al.'s \cite{akula2020cocox} report a user study that compared their own CoCoX counterfactual method to two other counterfactual methods (i.e., CEM \cite{dhurandhar2018explanations} and CVE \cite{goyal2019counterfactual}) showing that their method performed best, though  several methodological issues undermine these tests (see next subsection).  Finally, Kirfel \& Liefgreen \cite{kirfel2021whatifandhow} specifically tested for effects of ``actionable features" on counterfactual explanation (i.e., features that the user is known to be able to change).  They found that people’s perception of the quality and comprehensibility of explanations was affected the involvement of actionable and mutable features, as opposed to immutable ones. However, they also suggested that the actionable/mutability distinctions made in the methods were not clear-cut in the human domain and were often hard to define. So, in summary, the few comparative tests that have been carried out perhaps show less support for benchmark Min-Edit methods in favor of algorithms that are more concerned with computing within-domain explanations, though again the evidence is patchy.



\subsection{User Testing on Image Datasets}
\noindent Only a handful of papers consider user-tests of counterfacual explanations for image datasets, most focus  on tabular datasets. Yet, many counterfactual methods have specifically been proposed for image datasets (e.g.,\cite{goyal2019counterfactual,dhurandhar2018explanations, chang2018explaining, vermeire2022explainable, hendricks2018generating}).  Unfortunately, most of the few papers testing images have significant issues around their experimental designs, statistical analyses and/or the statistical significance of the results \cite{goyal2019counterfactual, larasati2020effect,  zhao2021generating, Singla2020Explanation}. This leaves us with three core papers that report anything indicative on the topic \cite{goyal2019counterfactual, cai2019effects, akula2020cocox}.

Goyal et al. \cite{goyal2019counterfactual} reported an influential method, Counterfactual Visual Explanation (CVE), that highlights key regions in an image (e.g., the beak colour of a bird) as feature differences behind counterfactual class changes (e.g., classifying a bird image as an auklet or a cormorant).   They performed a user study (N=26)\footnote{For appropriate statistical power this design requires an N$>$100 and confidence levels should be 95\% or higher.} with three conditions testing a no-explanation control against two explanation conditions (i.e., a non-counterfactual feature-region explanation and counterfactual-region explanation).  They found the counterfactual-region explanation elicited the highest accuracy (77.8\%), followed by the feature-region explanation (74.3\%), followed by the no-explanation controls (71.1\%), differences that were only significant at lower-than-usual confidence levels (i.e., 87\% and 51\%). So, at best, these results are indicative rather than conclusive.

Cai et al. \cite{cai2019effects} used QuickDraw Doodles (one of the datasets we use here) to reveal more conclusive results in a design that elicited better user interaction. They had participants (N=1150) generate  QuickDraw Doodles of common objects (e.g., draw a helicopter or an avocado) and then had a classifier identify the object using a dataset of labelled drawings.  The classifications produced were accompanied by \textit{normative explanations} (i.e., similar examples from the same class, such as other doodles of avocados) or \textit{comparative explanations} (i.e., counterfactuals or similar example from other classes, such as doodles of a pear or potato) with participants being asked to rate how well they understood the system and their views on the system's capability. The results showed that explanations only impacted misclassifications by the system (i.e., no effects for correct classifications) and that the example-based, normative explanations improved people's understanding and assessments of system capability.   Unfortunately, these effects did not extend to the counterfactual-based comparative explanations.  Cai et al. considered this failure to find counterfactual effects as being due to the ``surprisingness'' of the counterfactual examples, but the failure could also have been due to the presentation of items from multiple contrasting classes (i.e., people could be shown three different contrasting-classes instances). In passing, it should be said that this study is quite close to the current one, in that it pioneered the user-generation of QuickDraw images in an XAI user-study; however, it differs from the current study in that it asked users to generate training-instances rather than \textit{specifically} asking them to generate the counterfactual explanations for test-instances\footnote{To labor this point,  Cai et al.'s system retrieved the labelled-drawings of users and used them as counterfactual explanations. So, technically speaking, the counterfactuals were user-generated.  But it was their system that decided which of these drawings were used as counterfactual examples, rather than the participants in their study.  So, they were not user-generated explanations.}.

Finally, Akula et al.'s user tested their CoCoX method, which uses a ``fault-lines" technique that  feature-highlights counterfactual explanations akin to those advanced by CVE \cite{goyal2019counterfactual}.  CoCoX was compared against  CEM \cite{dhurandhar2018explanations} and CVE \cite{goyal2019counterfactual} alongside seven other non-counterfactuals methods\footnote{The training phase in these experiments appears show people the model's class-selection for 15 test-instances and then asks them for a counterfactual-class for that prediction, with no feedback.  This is an unusual training task that probably hampers transfer-of-training to non-counterfactual methods in the test phase of the study, perhaps explaining why these methods do so badly.} (e.g., LIME, GradCAM, CAM) using two measures: (i) a measure of people's agreement with the model's predictions for test-instances (they call this Justified Trust), and (ii) some of the satisfaction questions proposed by the DARPA group \cite{hoffman2018metrics}. The results showed that CoCoX does best on both measures with CEM and CVE competing for second position.  Furthermore, these explanation conditions do markedly better than no-explanation controls (i.e., ~30\%-40\% better). Although these authors are to be commended for their user-testing efforts, unfortunately this study has several serious design flaws. It appears to be designed as two separate 10-group between-subjects experiments, one for ML experts (N=20) and one for non-experts (N=60), neither of which are appropriately powered (i.e., a 10-condition experiment of this type would require several 100 participants). So, for instance, in the expert experiment this design means that the positive results found for CoCoX are based on just two participants seeing 5 test-items (i.e., 10 data-points), which could by-chance just happen to provide positive results. Furthermore, five test-materials seem too few; assuming all groups received the same five test-items, we do not know whether these particular test-instances just happened resonate well with training on CoCoX's explanations over other methods (i.e., a larger set of training and test-instances or randomly selected training- and test-instances should have been used).  So, again, these results on image-datasets are indicative rather than conclusive.

\subsection{Interim Conclusions on User Testing}
\noindent

In the previous sections, we have seen how the extant user-testing literature on counterfactual explanations has produced mixed and, sometimes, inconclusive results. Indeed, there are further recent results that seem to show that other exogenous factors (i.e., variables independent of counterfactual manipulations) have greater impacts; factors such as (i) expertise or familiarity with the domain \cite{ford2022}, (ii) whether explanation tasks are framed as predictions or diagnoses \cite{dai2022counterfactual}, and (iii) whether instances involve categorical or continuous features \cite{Warren2022FeaturesXAI_IJCAI,warren2022better}.    To us, these findings suggest that the community may not be taking the right approach to users, that current user studies has not sufficiently human-centered.  Also, that taking such a perspective may shed a very different light on our current understanding of how people understand counterfactual explanation.  In the next section, we consider a new methodology which attempts to rectify these failings.

\begin{figure}
\centering
\begin{subfigure}[b]{1\textwidth}
   \includegraphics[width=1\linewidth]{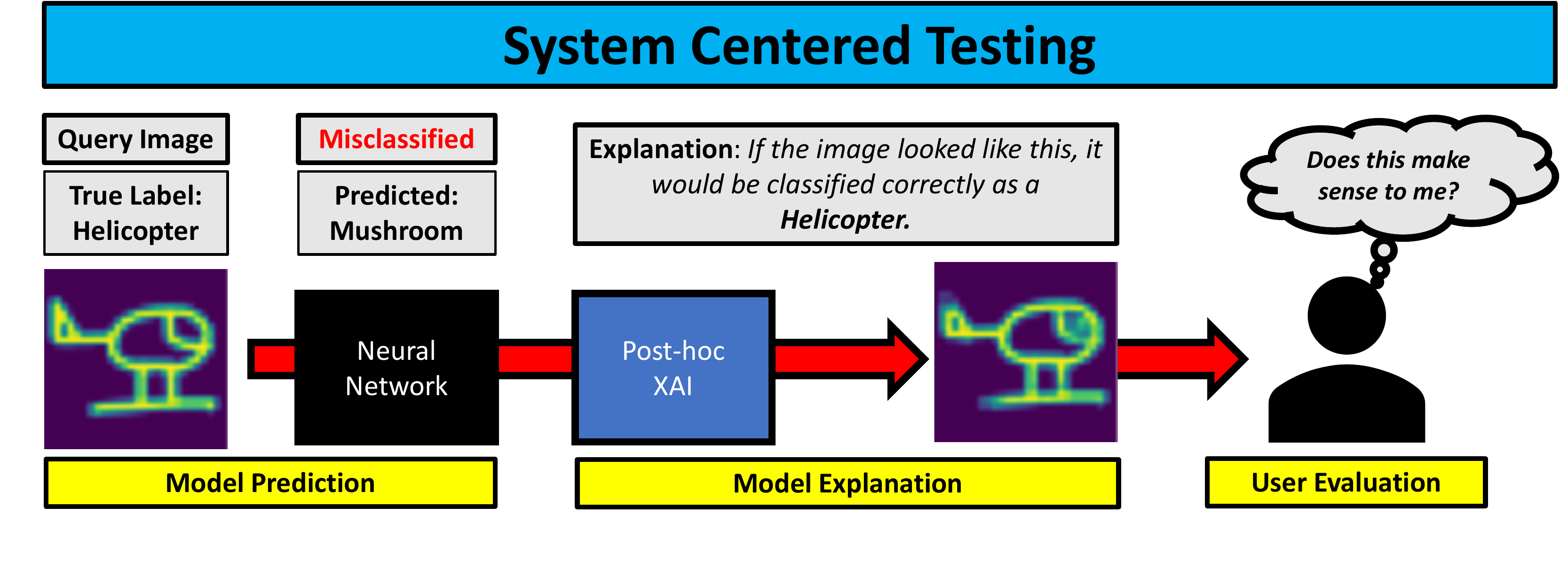}
   \label{fig:Ng1}
\end{subfigure}
\begin{subfigure}[b]{1\textwidth}
   \includegraphics[width=1\linewidth]{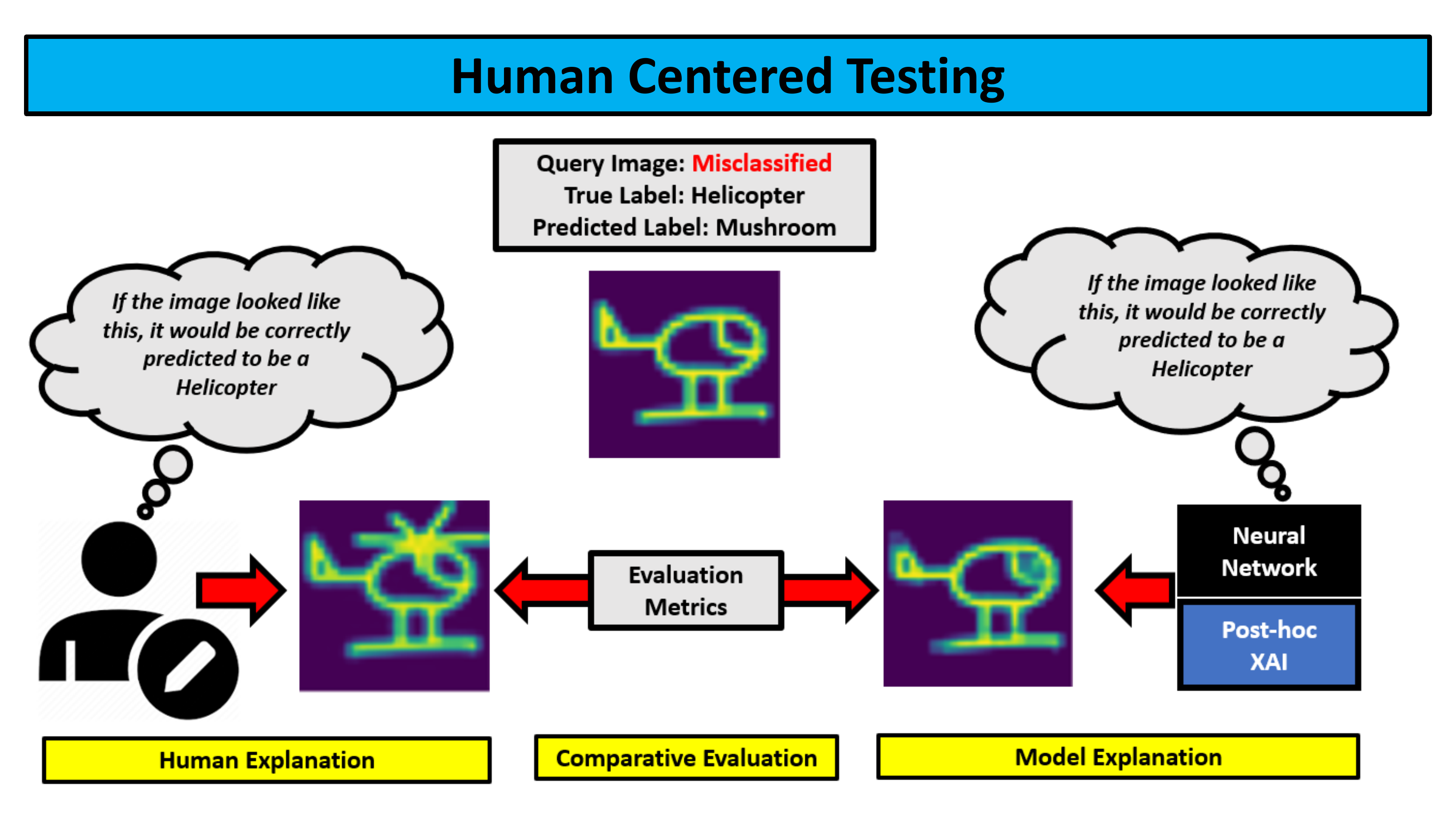}
   \label{fig:Ng2}
\end{subfigure}
\caption[]{System-centered user-tests of counterfactual XAI present people with the outputs from an AI-model and XAI method to evaluate them in different ways.  User-centered tests try to capture the user's perspective on explanation. Here an XAI-method's explanation of model outputs (e.g., misclassified images) are evaluated by comparing them to human explanations of the same model outputs (e.g., misclassifications) which are collected as a ground-truth. Most, if not all, of the current user-tests of counterfactual XAI are system- rather than user-centred.}
\label{fig:system_vs_human}
\end{figure}

\section{A User-Centered Methodology for Counterfactual XAI}
\label{user_centered}

\noindent All the user studies that have been reviewed thus far have been decidedly \textit{system-centered} ones, in which users are cast as passive recipients of machine-generated explanations.  In these studies, XAI methods are used to generate explanations for AI-model outputs, that are then fed to people to be evaluated in different tasks (e.g., for correctness, acceptability, helpfulness, trustworthiness).  Such studies lack a reality-check on whether these machine-generated explanations are actually the ones that people really require; for instance, we do not even know if these machine-explanations are  representative of human counterfactual explanations. In contrast, a more \textit{user-centered} approach would focus on the user, on \textit{their} conception of counterfactual explanation.  So, arguably, the user-testing of counterfactual XAI requires a Copernican reversal from being overly system-centered to being more user-centered (see e.g., Figure \ref{fig:system_vs_human}).
However, to date, the XAI community has clearly found it hard to design such user-centered studies\footnote{In many ways, \cite{Kaushik2020Learning} is very different to the current work in its focus on data augmentation in NLP, but conceptually it tries to be user-centered in the same way, as it uses a human-in-the-loop task to elicit people's counterfactual understanding of sentiment in texts}.  Here, we advance a novel user-centered methodology that starts with gathering a user-generated ground-truth of counterfactual explanations for a CNN's misclassifications, before comparing these human-generated explanations to the machine-generated ones from several benchmark counterfactual methods.   This user-centred study allows us to verify the plausibility claims made for different counterfactual XAI algorithms.  In the following sections, we introduce (i) this new two-step methodology, (ii) the editing tool used to collect the ground-truth explanations, (iii) the evaluation metrics used to compare the human-generated and machine-generated explanations.


\subsection{A User-Centered Two-Step Methodology}
\label{sec:usercentered}
\noindent The current methodology realises a human-centered approach in two steps: (i) ground truth data collection, followed by (ii) comparative evaluations of human- and machine-generated explanations. Two datasets are used: the benchmark MNIST images of written Arabic numbers \cite{lecun1998mnist} and QuickDraw Doodle images  \cite{cai2019effects}.  The QuickDraw dataset is arguably more  complex than the MNIST one; notably, it involves images with parts that people can readily name (e.g., the toppings on a pizza slice). We train CNNs for each of these datasets and randomly select a sample of misclassifications made by the models.  These misclassified instances are then presented to (i) human participants in a psychological experiment and (ii) to each counterfactual method to collect the explanations generated.  The two main steps in the methodology are as follows:
\begin{itemize}
\item \textit{Ground-Truth Collection.} People were provided with a simple editing tool to create their own counterfactual explanations for misclassified images from the CNN, to collect ``ground-truth explanations" for each dataset. This  collection was done in two separate experiments, one using the MNIST items (N=42) and a separate pilot using the QuickDraw items (N=5).

\item \textit{Human-Machine Comparative Evaluation}. The same misclassified images were then presented to each of four counterfactual methods -- Min-Edit, CEM-PN, VLK and Revise -- to produce parallel sets of machine-generated explanations, before doing a human-machine comparative evaluation of the explanation sets. We use benchmark evaluation metrics that have previously been used in computational evaluations of counterfactual methods to assess plausibility claims (i.e., on proximity, representativeness, prototypicality).
\end{itemize}
\noindent This methodology definitively tests whether the explanations generated by people correspond in any way to those generated by these counterfactual methods. As such, to the best of our knowledge, this is the first true user-centered assessment of counterfactual algorithms.  In the following sub-sections, we provide more details on each of the steps in this methodology.

\begin{figure}[h!]
\centering
\includegraphics[scale=0.5]{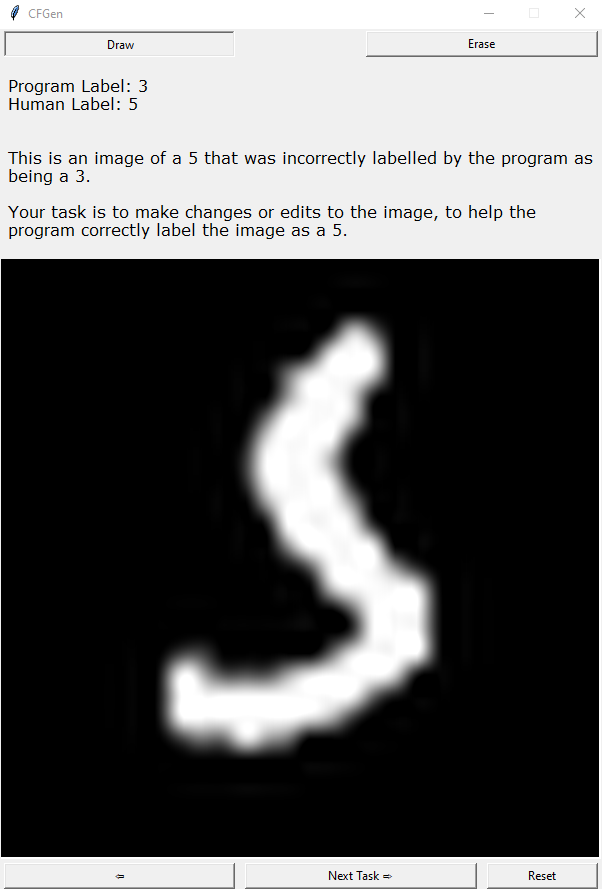}
\caption{Screenshot of the editing tool used for collecting ground-truth explanations, showing a misclassified MNIST image of a ``5", along with the instructions to participants. The interface allows pixels to be added or removed using the cursor as a pen or eraser, after clicking the ``Draw" or ``Erase" buttons, respectively. The ``Reset" button removes all edits, resetting the image to its original form.}
\label{fig:screenshot}
\end{figure}

\subsection{The Ground-Truth Step: Task \& Tool}
\noindent
For the first step in the methodology, a simple software tool was developed to present the misclassified images to participants.  The tool allowed images to be edited via a custom interactive GUI implemented using the tkinter Python package (see Figure \ref{fig:screenshot}).  The ground-truth collection was carried out by presenting the CNN's misclassifications to people and asking them to edit the query-image to correct the incorrect prediction.   For each misclassification, they were told the model's label and its correct label (e.g., 3 and 5, respectively) and that it was misclassified (i.e., ``This is an image of a 5 that was incorrectly labelled by the program as being a 3"). They were then invited to edit the image using the editing tool, to explain how the misclassification would have to change to be correctly labelled (i.e., ``Your task is to make changes or edits to the image, to help the program correctly label the image as a 5"). This task requires people to create a counterfactual instance that shows how the image would have to change to be correctly classified.

In the user test involving the MNIST dataset, two separate groups of participants were given slightly different instructions, the so-called ``Normal" and ``Min-Edit" groups.  The Normal group was given the instructions discussed above, asking participants to  ``\textit{...make changes or edits to help the program correctly label the image...}". In contrast, the Min-Edit Group was asked to ``\textit{...make the smallest possible changes needed, to help the program correctly label the image...}''.  As we saw in Section \ref{Related Work I}, a key assumption of many algorithms is that the counterfactual should make minimal changes or ``minimal edits" to the query. This instructional manipulation was designed to determine whether instructions to uses to act in accordance with the Min-Edit method changed responding relative to the ``normal" non-directive instructions.

\subsection{The Comparative-Evaluation Step: Three Perspectives}
\label{sec:comparative_evaluations}

\noindent For the second step in the methodology, the comparative-evaluation step, the counterfactual explanations produced by human and machine were systematically compared using key evaluation metrics that are commonly used in this area.  This section describes each of the evaluation metrics used for these comparisons grouped by (i) \textit{proximity} tests comparing distances between query and counterfactual instances (using L1 and L2 norms), (ii) \textit{representativeness} tests assessing generated counterfactual instances (using Monte Carlo Dropout, IM1, 10-LOF, and R\% Sub, see Section \ref{Representativeness Metrics}), and (iii) \textit{prototypicality} tests comparing distances generated counterfactual instances to class prototypes (using Grad-Cos on prototypes retrieved from MMD-Critic).  Taken together, these evaluations provide a definitive test of the extent to which current XAI methods produce plausible, counterfactual explanations and the extent to which they diverge from those produced by people.  In the following subsections, we describe the specific metrics used.

\subsubsection{Proximity Metrics: L1 \& L2 Norms}
\noindent The distance metrics -- \textbf{L1} (Eqn. \ref{l1}) and \textbf{L2} (Eqn. \ref{l2}) norms -- have typically been used to evaluate counterfactual methods, measuring the closeness of the counterfactual image, $I'$,  to the query image, $I$,  where lower distance-scores are assumed to be a proxy for explanation quality and plausibility (but see \cite{keane2021if} for criticism). Here, we compare distance scores for machine-generated query-counterfactual pairs to the corresponding ground-truth pairs produced by people in two user studies (see the MNIST and QuickDraw experiments in Sections \ref{sec:setup} and \ref{sec:procedure}).



\begin{equation}
\label{l1}
\sum_{i} | I_{i} - I'_{i} |
\end{equation}

\begin{equation}
\label{l2}
\sqrt{(\sum_{i} |(I_{i} - I'_{i})|^{2})}
\end{equation}


\subsubsection{Representativeness Metrics: Out-of-Distribution (OOD) Measures}
\label{Representativeness Metrics}
 \noindent Point representativeness and out-of-distribution metrics -- Monte Carlo Dropout \cite{gal2016dropout, kenny2020generating, delaney2021uncertainty}, R\%-Substitutability \citep{samangouei2018explaingan, kenny2020generating}, LOF \cite{breunig2000lof, kanamori2020dace, delaney2021instance}, IM1 \cite{VanLooveren2019, kenny2020generating} -- have often been used to evaluate  counterfactual methods. These measure the representativeness of explanations to the data-distribution or classes found in the distribution. Here, we briefly describe these metrics that are used to compare the machine-generated counterfactuals and those generated by people.

\paragraph{\textbf{Monte Carlo Dropout}}
Originally proposed by \cite{gal2016dropout}, MC-Dropout enables one to quickly estimate the uncertainty associated with a prediction by inspecting the variance of a predictive distribution \cite{AWS_Uncertainty}. Multiple stochastic forward passes with different dropout configurations can yield this predictive distribution. Following \cite{kenny2020generating, delaney2021uncertainty} we leverage MC-dropout to evaluate counterfactual explanations by estimating the posterior mean of the predictive distribution \textbf{MC-Mean} (higher is better) and the posterior standard deviation \textbf{MC-Std} (lower is better). The intuition is that explanations with lower uncertainty scores should be more representative of the counterfactual class as they are better grounded in the data distribution. Full technical details of MC-Dropout and the implementation can be found in Appendix A.

\paragraph{\textbf{R\%-Substitutability}}
Inspired by \cite{samangouei2018explaingan, kenny2020generating}, the generated counterfactuals are used as training data to fit to a $k$-NN classifier (in pixel space) which then predicts the full test-set. For MNIST, as we are using 50 instances we compare to an MMD Prototype 1-NN classifier \cite{kim2016criticism} that achieves 75.57\% accuracy on the full MNIST test set (10,000 images), using only 50 prototypical instances and a Euclidean distance function. A method that achieves half this accuracy would achieve an R\% - Substitutability score of 50\%.

\paragraph{\textbf{IM1}}
Originally presented by \cite{VanLooveren2019} as an interpretability metric, a convolutional autoencoder is trained on the predicted class $c$, $(AE_{c})$ and on the counterfactual class $c'$, $(AE_{c'})$ and the $l_{2}$ reconstruction error is monitored to compute IM1. Specifically:
\begin{equation}
    IM1= \frac{\||I'-AE_{c'}(I')\||_{2}^{2}}{\||I'-AE_{c}(I')\||_{2}^{2}}
\end{equation}
A lower value of IM1 implies that the candidate counterfactual image $I'$ can be better reconstructed by autoencoders that have seen instances of the counterfactual class, relative to an autoencoder that has seen instances in the original class, implying that $I'$ lies closer to the data manifold of $c'$.

\paragraph{\textbf{10-LOF}}
Originally presented by \cite{breunig2000lof}, the local outlier factor (LOF) algorithm is a distance-based technique that determines whether an instance is out of distribution by computing the local density deviation with respect to its neighbours in the pixel space. Following \cite{kanamori2020dace}, the 10-LOF can be used to determine if a counterfactual explanation is plausible according to the data distribution. The decision-score metric is centred on zero, with higher values indicating that a sample is more within the distribution according to 10-LOF.

\subsubsection{Prototypicality Metric: MMD-Critic \& Grad-Cos}
\label{prototypicality_metric}
 \noindent MMD-critic \cite{kim2016criticism} is implemented to compute prototypes by minimizing the maximum mean discrepancy between the prototype distribution and the data distribution using a kernel density function, full technical details can be found in Appendix B. This evaluation aims to determine whether generated counterfactuals are actually close to prototypes for the counterfactual class.  Recall, Min-Edit methods try to find instances that are close to the query but just over the decision boundary in a contrasting class; hence, they are much less likely to be representative members of this counterfactual class. The Grad-Cos metric allows us to determine whether machine-generated and human-generated counterfactuals are close to prototypes and is briefly described below.

\paragraph{\textbf{Latent Space Similarity: Grad-Cos}}
Given some labelled input image $I_{A} = (x,y)$ and a black-box neural network, $b_{\theta}(I)$ that is parameterized by $\theta$, with loss $\ell(I; \theta)$ and gradient $\nabla_{\theta}\ell(I;\theta)$;
Grad-Cos \cite{charpiat2019input} is a gradient based similarity metric that quantifies the degree to which the loss will change when a small update to the model is made using some candidate training instance, $I_{B}$, (e.g., a class prototype). If these two images are very similar from the neural network's perspective, this change will be large. Formally, the cosine similarity of gradients kernel can be expressed as:
\begin{equation}
   k_\theta(I_{A},I_{B})  = \frac{{\nabla_{\theta}\ell(I_{A}) \cdot \nabla_{\theta}\ell(I_{B})}}{\||\nabla_{\theta}\ell(I_{A})\||\||\nabla_{\theta}\ell(I_{B})\||}
\end{equation}
Recent research independently evaluated several popular relevance metrics (e.g., Influence Functions \cite{koh2017understanding}) and found Grad-Cos to be the recommended choice in practice \cite{hanawa2021evaluation}. Grad-Cos passed the weight randomization tests of \cite{adebayo2018sanity}, indicating that it is faithful to the underlying machine learning model.

\section{A Comparative Study: Human \& Machine Explanations}
\label{sec:user_study_ground_truth}
\noindent
As described in Section 4, a two-step, user-centered methodology was applied to the explanation of a CNN's misclassifications after training on two datasets (the MNIST and QuickDraw Doodles datasets).  Two user experiments, one for each dataset, were carried out to gather ground-truth explanations for each of the selected misclassifications. Machine-generated explanations were generated for each of the same misclassifications using four benchmark counterfactual XAI methods -- Min-Edit, CEM-PN, VLK and Revise (see \ref{Related Work I}). These user-generated and machine-generated explanations where then compared using benchmark evaluation metrics for proximity, representativeness and prototypicality.   The following sub-sections lay out the method for this overarching comparative study and specific methods used in the two user-tests that collected the ground-truth data.

\subsection{\textbf{Method: Comparative Study}}
\label{sec:setup}
\subsubsection{Model Setup: CNN Classifier \& Datasets}
\noindent  The to-be-explained, black-box model was a convolutional neural network (CNN) trained using a well-known architecture  \cite{VanLooveren2019}. Two image datasets were used: the MNIST \cite{lecun1998mnist} and Google QuickDraw \cite{cai2019effects} datasets. The  \textit{MNIST dataset} contains images of written numbers, with 70,000 images covering 10 classes (i.e., the digits 0--9). These images were scaled to $[-0.5,0.5]$ and the default training and test sets were used. Dropout layers were implemented for regularization and to facilitate uncertainty computations, using MC-Dropout \cite{gal2016dropout}. The CNN was trained with an Adam optimiser for 10 epochs using a batch size of 256, to achieve an accuracy of 98.93\% on the test set, resulting  in 107 to-be-explained images being misclassified by the model. Pretrained model weights are provided to aid reproducibility.  The Google QuickDraw dataset contains images gathered from studies that presented people with common objects, asking them to draw the object in 20 seconds as a ``doodle''. It has 50 million doodle images covering 345 classes (i.e., common object categories such as ``bicycle'' or ``helicopter''). The architecture of the classifier was the same as that used on the MNIST dataset. This CNN was trained on a sample of 35,000 images from 5 categories (i.e., ``bicycle'', ``giraffe'', ``helicopter'', ``mushroom'', and ``pizza'') using an Adam optimiser for 10 epochs with a batch size of 256. The model achieved an accuracy of 97.02\% on the test set, resulting in 447 to-be-explained images which were misclassified.

\subsubsection{Materials: Comparative Study}
\label{sec:study_materials}
\noindent The same misclassifications were presented to people in the user-tests and to the counterfactual methods for the comparative study.  For the MNIST dataset, 50 misclassified images were randomly selected from 107 MNIST images misclassified by the CNN.   For the QuickDraw dataset, 30 misclassified QuickDraw Doodles were randomly selected from 447 QuickDraw images misclassified by the CNN.

\subsubsection{Explanation Methods: Comparative Study}
\label{sec:study_methods}
\noindent Four state-of-the-art counterfactual methods were selected from the literature on counterfactual XAI \cite{karimi2020algorithmiccf,keane2021if} based on their (i) popularity as benchmark methods (i.e., according to citations), (ii) their availability as maintained open-sourced code (e.g., on GitHub), and (iii) their ability to handle image data. The selected benchmark methods, previously outlined in Section \ref{Related Work I},  are: Min-Edit \cite{wachter2017counterfactual}, CEM \cite{dhurandhar2018explanations}, VLK \cite{VanLooveren2019} and Revise \cite{joshi2019towards}.

\subsubsection{Procedure: Comparative Study}
\label{sec:study_procedure}
\noindent The two user tests were run as independent experiments on each of the datasets, one for MNIST and one for the QuickDraw data. The responses from these user studies were post-processed to find the mean group responses for each misclassified item. The same misclassifications were presented to the four counterfactuals methods and the generated explanations recorded. Pairwise comparisons between the human explanation for a given misclassification and the method-generated explanation for that misclassification made by each of the four counterfactual methods.
These pairings were then evaluated with respect to the six metrics for assessing different aspects of plausibility (see next subsection).  Finally, statistical summaries of the results from the metrics were collated and analysed using statistical tests where appropriate.

\subsubsection{Evaluation Measures: Comparative Study}
\label{sec:study_procedure}
\noindent Six evaluation metrics were used to process the pairings of human and machine explanations (for detailed descriptions see Section \ref{sec:comparative_evaluations}).  The metrics were (i)  L1, (ii) L2, (iii) MC-Mean, (iv) MC-Std, (v) IMI, (vi) 10-LOF, (vii) R\% Sub and (viii) Grad-Cos Similarity.

\subsection{\textbf{Method: User Tests}}
\subsubsection{Participants and Design: User Tests}
\noindent Forty-seven participants took part in the two user studies: the MNIST Study (N=42) and QuickDraw Study (N=5). In the MNIST Study, participants were randomly assigned to two independent groups, the Normal and Min-Edit Groups (both N=21).  This sample size was based on a power analysis designed to balance the probability of Type I and Type II errors. Using GPOWER \cite{erdfelder1996gpower}, for a two separate one-way, t-tests design, with the assumption of a large effect size for each (d = .0.8), the power analysis showed that an N of 42 ensured an alpha of .05 and power of .80. The QuickDraw Study was designed as single-group pilot to determine whether the findings for the MNIST data generalised to the, arguably, more-complex QuickDraw dataset. Both studies were reviewed by the university's ethics board (ref. LS-E-21-215-Delaney-Keane). Participants were Computer Science students at UCD and were paid an hourly rate of €13.00 in accordance with the living wage in the jurisdiction.

\subsubsection{Apparatus: User Tests}
\label{sec:software}
\noindent The software tool was developed that allowed images to be edited via a
custom interactive GUI implemented using the tkinter Python package (see Figure \ref{fig:screenshot} for a screenshot). The presented image was up-sampled to a $600 \times 600$ canvas where it could be edited and the final image was down-sampled to the original $28 \times 28$ size. Participants had the option to add pixels, remove pixels or reset the image to its original form if they made a mistake. A log of the stroke information carried out by the user and the final edited image for each presented image was recorded and saved for later analysis.

\subsubsection{Procedure: User Tests}
\label{sec:procedure}
\noindent In both of the user experiments, all participants were tested in a face-to-face experiment with a single experimenter (ED) who presented them with task instructions and the editing tool. After reading the instructions, participants were given three practice trials to learn how to use the tool.  The software tool presented one misclassified image at a time, along with (i) a statement about the label it was given by the program and its correct label given by humans, (ii) a statement about how the image was incorrectly labelled by the program, and (iii) the explanation-task instructions to edit the image to help the program correctly label the image. This task asks the user to counterfactually explain the program's misclassification to help improve its subsequent learning. In the MNIST Study, the Normal Group were instructed to ``\textit{...make changes or edits to the image, to help the program correctly label the image...}", whereas the Min-Edit Group were instructed to ``\textit{...make the \underline{smallest possible changes} needed, to help the program correctly label the image...}". The latter instructions encouraged this group  to conform to a key assumption made by many XAI methods (e.g., the Min-Edit assumption), even though this requirement seems somewhat artificial in this context. In the QuickDraw Study, all participants were given the Normal-Group instructions to ``\textit{...make changes or edits to the image, to help the program correctly label the image...}". In both studies, after receiving the instructions and practice trials, participants proceeded through all the presented images at their own pace. The presented set of images was randomly shuffled anew for each participant to control for possible order effects.   Each experimental session took $\sim$15-30 minutes (typically, $\sim$20 min in MNIST Study and $\sim$15 min in QuickDraw Study), including the final de-briefing on the rationale for the study.  The logs of participant's stroke information and final edited image for each item were all recorded and saved after being suitably anonymised.

\subsubsection{Response Post-Processing: User Tests}
\label{sec:procedure}
\noindent In each of the user studies, for a given misclassified item a response from each participant in the experiment is recorded; so, for MNIST experiment we have 42 explanations for the first misclassification, 42 for the second and so on.  So, overall 2,250 human explanations were gathered: 42 people x 50 items for the MNIST experiment and 5 people x 30 items for the QuickDraw experiment.  However, for each of the counterfactual methods we have just one explanation  per misclassification; so, 320 explanations (80 items x 4 methods).  So, to compare the human and machine explanations in a one-to-one fashion, we computed the centroid of human responses to a given item.  This group-level response was then used in the explanation-to-explanation comparison for each of the metrics. 

\subsection{\textbf{Results \& Discussion: Comparative Study}}
\label{sec2}

\begin{figure}[h]
\centering
\includegraphics[width=0.86\linewidth]{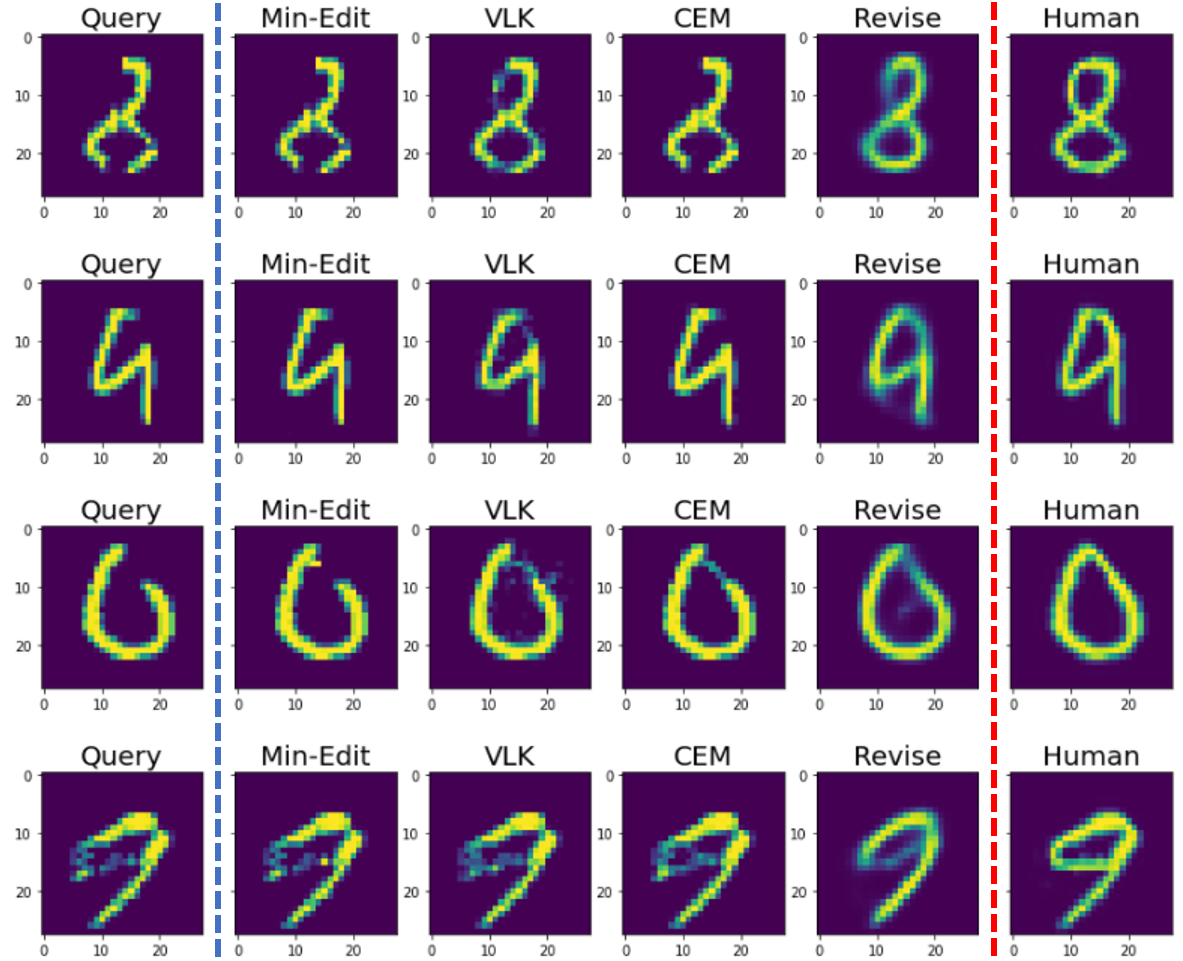}
\caption{Using the MNIST dataset, four misclassified query images and their corresponding counterfactual explanations generated by four XAI methods (Min-Edit, VLK, CEM-PN and Revise) and by humans (from the MNIST User Study).}
\label{fig:mnist}
\end{figure}

Figure \ref{fig:mnist} shows some representative data on the types of explanations generated by people and the fours methods examined; even a cursory glance at these items shows that the human explanations tend to be more complete and identifiable instances of the counterfactual class for the misclassified number.  These explanations collected from the two user-tests and those generated by each of the four methods were comparatively analysed using the six evaluation measures.  The evaluation results are grouped and reported under the headings of proximity, representativeness and prototypicality. Significantly, these evaluations do \textit{not} support the intuition-based claims made for many current counterfactual methods.   In summary, they show: 

\vbox{
\begin{itemize}
\item \textit{Proximity Evaluations}: the distance measures for machine-generated query-explanation pairs were significantly closer to one another than the ground truth, human-generated pairs; human counterfactual explanations are \textit{not} Min-Edits of queries, instead humans make large edits to the query when generating counterfactuals (see Section \ref{sec:proxeval}).
\item \textit{Representativeness Evaluations}: representativeness tests reveal that people's counterfactuals are much more representative of the counterfactual class than the machine-generated explanations; indeed, they are further from the decision boundary and closer to the centre of the counterfactual class distribution (see  Section \ref{sec:repeval}).
\item \textit{Prototype Evaluations}: human counterfactual explanations are much closer to prototypes, computed in the latent space; they tend to sparsely modify high-level semantic features, in ways that counterfactual methods do not mimic well (see Section \ref{sec:proto}).
\end{itemize}}

\noindent In the following sub-sections we elaborate on the detailed evidence for these summary findings.

\subsection{Proximity: Human Counterfactuals are Not Min-Edits}
\label{sec:proxeval}

\begin{figure}[h!]
\centering
\begin{subfigure}{.5\textwidth}
  \centering
  \includegraphics[width=1\linewidth]{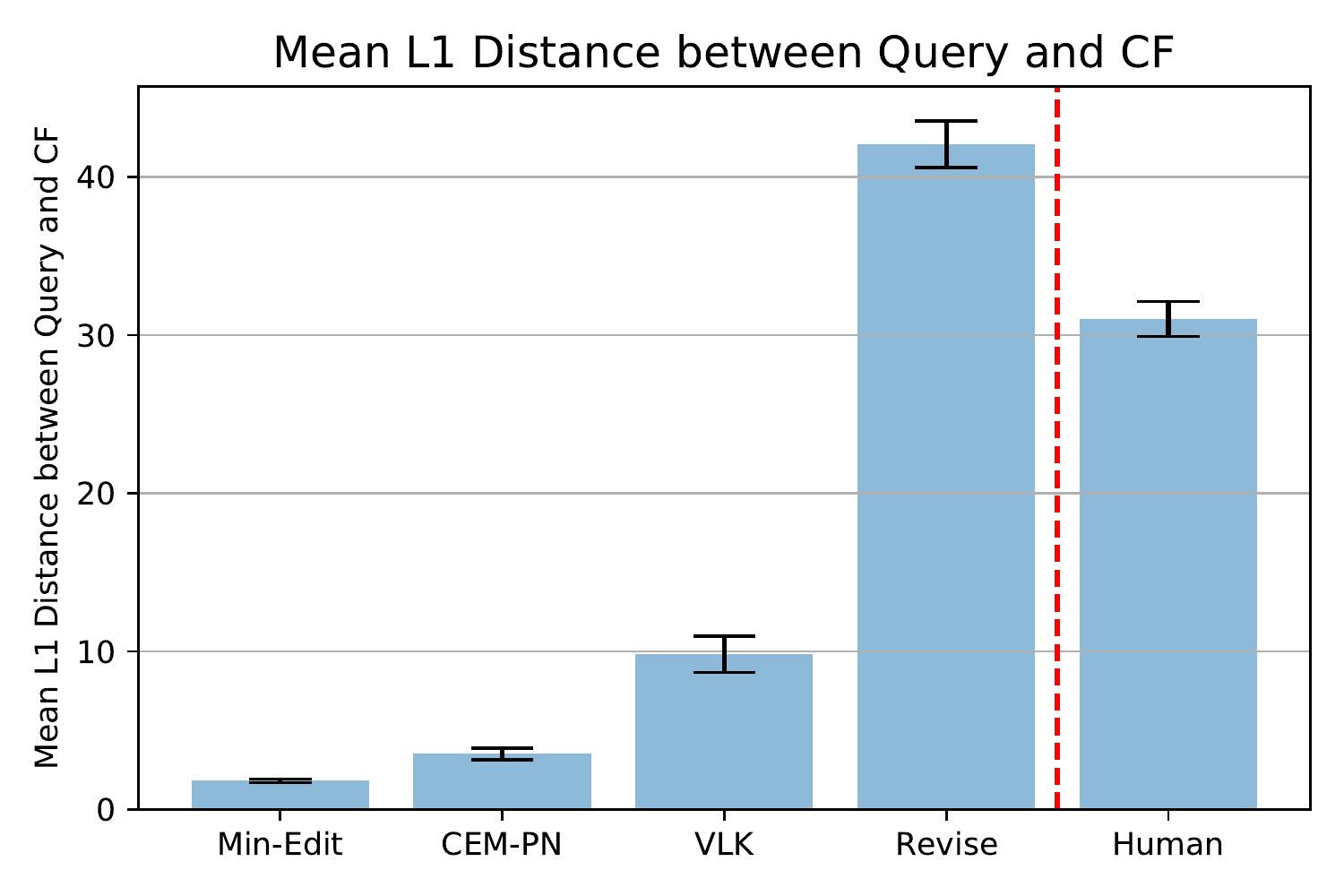}
  \caption{}
  \label{fig:sub1}
\end{subfigure}%
\begin{subfigure}{.5\textwidth}
  \centering
  \includegraphics[width=1\linewidth]{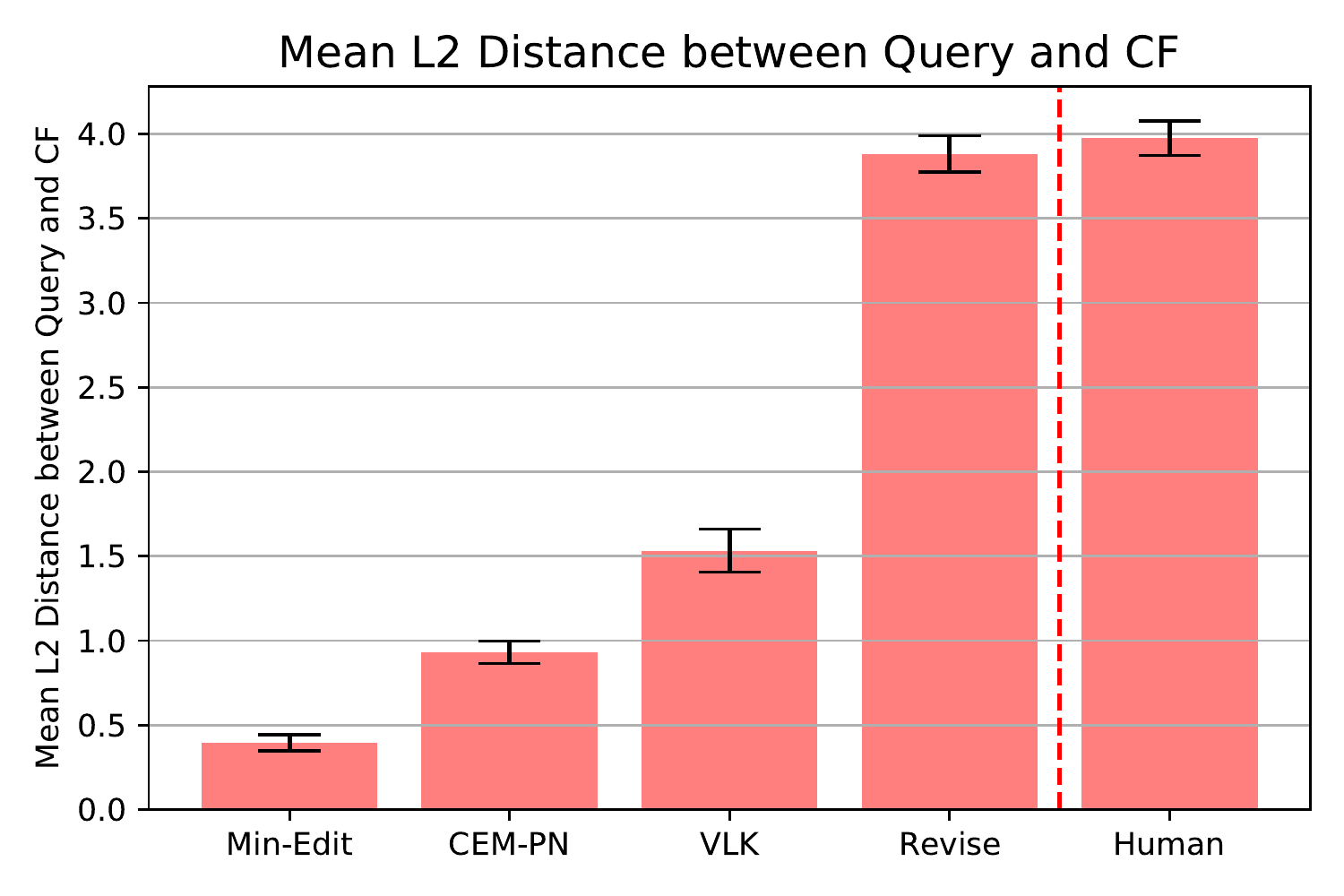}
  \caption{}
  \label{fig:sub2}
\end{subfigure}
\begin{subfigure}{.5\textwidth}
  \centering
  \includegraphics[width=1\linewidth]{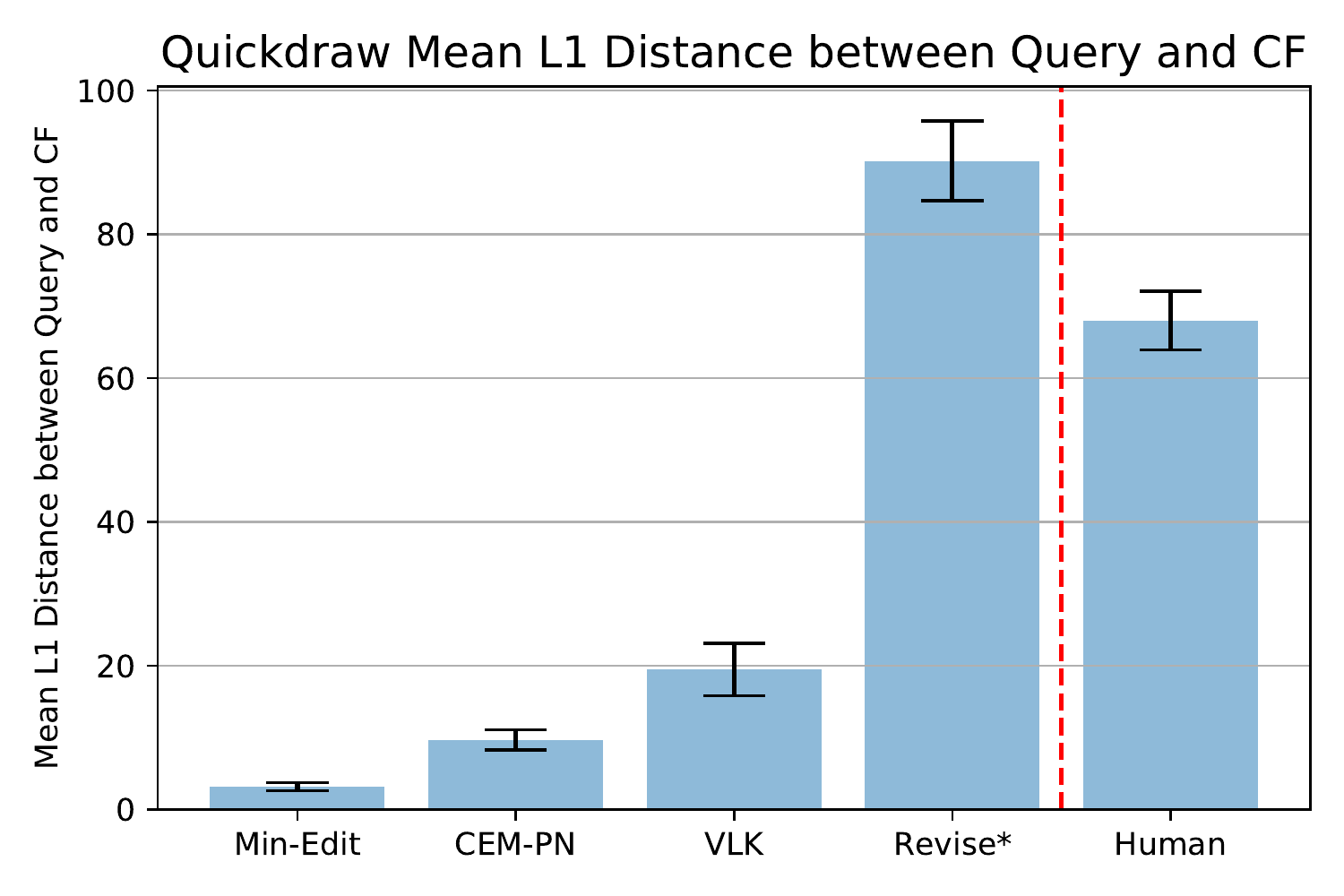}
  \caption{}
  \label{fig:sub1}
\end{subfigure}%
\begin{subfigure}{.5\textwidth}
  \centering
  \includegraphics[width=1\linewidth]{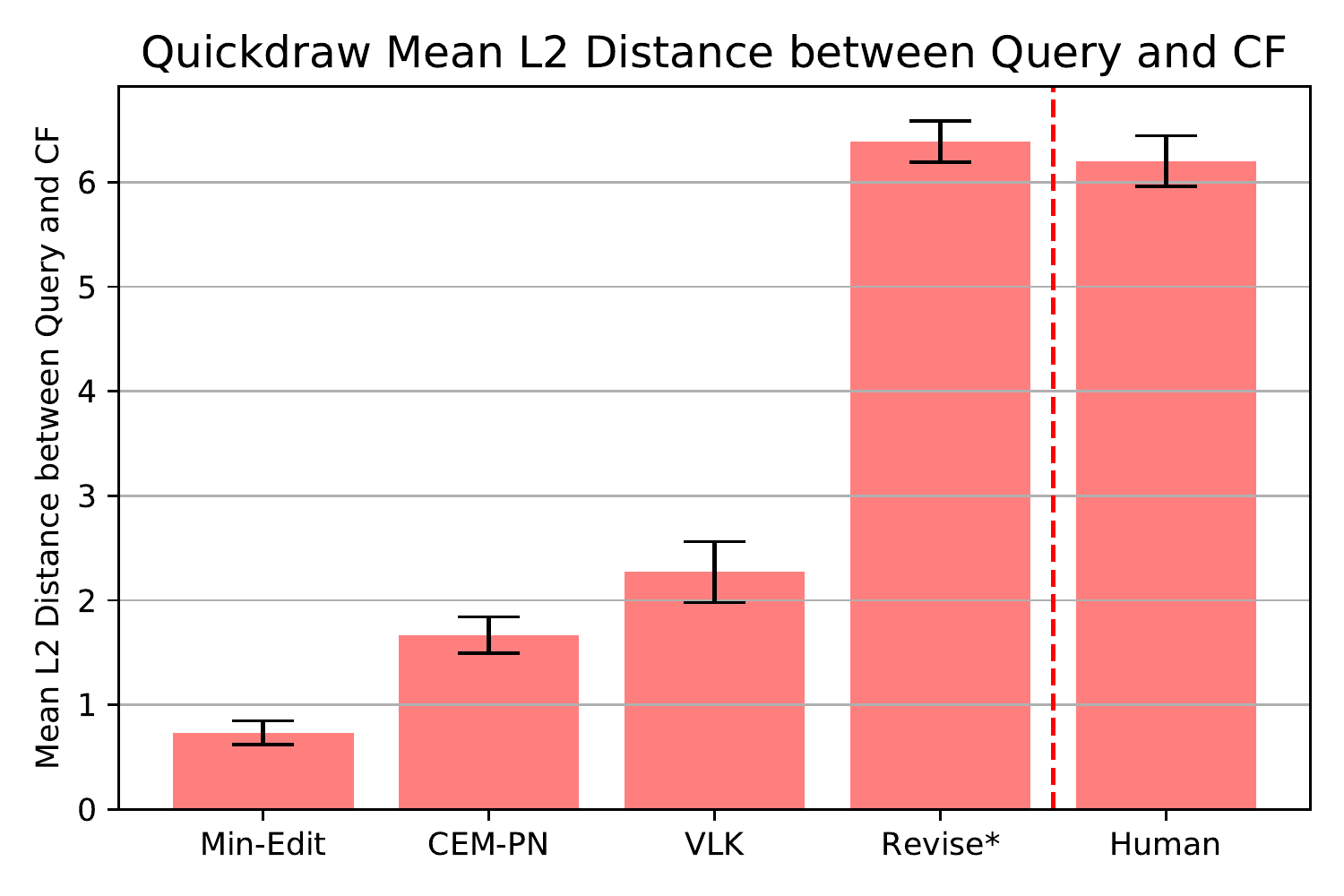}
  \caption{}
  \label{fig:sub2}
\end{subfigure}
\caption{\textit{Proximity Evaluations}. Mean L1 and L2 distance scores for query-explanation pairs produced by four counterfactual XAI methods -- Min-Edit, CEM-PN, VLK, Revise -- (left of the dotted red line) compared to the human ground-truth (right of dotted red line), for (a, b) MNIST and (c, d) QuickDraw datasets (Error bars show standard error of the mean).
*Note, Revise failed to generate counterfactuals for many instances in the QuickDraw dataset (a coverage deficit also found by \cite{holtgen2021deduce}); so, Revise's results only reflect instances where explanations were found, making its quite poor performance look better.}
\label{fig:proxeval}
\end{figure}

\noindent Popular proximity-based counterfactual methods aim to produce Min-Edit counterfactuals that minimise the distance between the query and explanation instances, achieving a class change between their predictions \cite{wachter2017counterfactual, wexler2019if,Mothilal2020,VanLooveren2019,dandl2020multi}. Hence, proponents of this view argue that plausible counterfactual explanations are ones that deliver lower L1 or L2 distances between the query and counterfactual.  However, this is a proxy evaluation in the absence of a human ground-truth; here, counterfactual goodness can be directly tested by comparing query-explanation distance profiles to human ones.  Figure \ref{fig:proxeval} shows that the distance profiles using L1 and L2 norms for the automated methods, all differ from the human profiles in tests using the MNIST and QuickDraw data.

\textit{For the MNIST data}, a statistical analysis, using a one-way ANOVA, of the distance metrics found a reliable main effect of Group for L1, $F$(4,45)$=$ 331.81, $p<0.001$, and L2,  $F(4,45) = 314.14$, $p<0.001$ (see Figure 4a and 4b).  Pairwise comparisons between the groups shows that the L1 and L2 scores for three methods (Min-Edit, CEM-PN, VLK) were all significantly lower that those for humans (all $p < 0.001$; using t-tests and a Bonferroni-Holm correction). In contrast, the Revise method is much closer to the human ground-truth; on L1 its distance scores are higher than human ones ($p < 0.001$) but on L2 it is not reliably different ($p > .05$).

Figure \ref{fig:mnist} provides a representative sample of data from the MNIST user-test, revealing how the Revise and human counterfactuals tend to be well-formed examples of the counterfactual class (intuitively, they appear almost prototypical). In contrast, the other three methods produce counterfactuals that are minimal changes to the original query, that are not representative members of the counterfactual class. In short, when people generate counterfactuals they do not produce counterfactual explanations that are Min-Edits of the query, contradicting a key assumption of most popular XAI methods.

Notably, this pattern of results does not change, even when participants were specifically instructed to use a Min-Edit strategy. Recall, in the MNIST user-test, half of the participants were specifically instructed to minimally-edit the images in contrast to more ``normal" instructions (Min-Edit versus Normal groups).  Although, these instructions significantly reduced the distance scores in the Min-Edit group relative to the Normal group (t-tests, all $p < 0.001$), the human distance scores for query-explanation pairs were all still significantly different and higher than the machine-generated ones (see Appendix C for details).  So, even when we instruct people to act in a Min-Edit way, they do not Min-Edit the images to the same degree as the methods do. 

\textit{For the QuickDraw data}, the L1 and L2 distance in the pixel space, show essentially the same patterns between groups; a one-way ANOVA found a reliable main effect of Group for L1, $F$(3,26) $=$ 107.03, $p<0.001$, and L2,  $F(3,26) = 123.62$, $p<0.001$ (see Figure 4c and 4d).  However, as we shall see later, exploring these distances in the latent space may more appropriate and informative. 

 \begin{table}[ht]
\centering
\caption{\textit{Representativeness Evaluations}. Four out-of-distribution measures for the XAI methods (Min-Edit, VLK, CEM-PN and Revise) compared to human responses for A - MNIST and B - QuickDraw (bold indicates best score in each case).}
\resizebox{\linewidth}{!}{
\begin{tabular}{@{\extracolsep{1.5pt}}lcccccccccc}
\toprule
{\textbf{}} & \multicolumn{2}{c}{MC-Mean} & \multicolumn{2}{c}{MC-Std} & \multicolumn{2}{c}{IM1} & \multicolumn{2}{c}{10-LOF} & \multicolumn{2}{c}{R\%-Sub}\\
 \cmidrule{2-3}
 \cmidrule{4-5}
 \cmidrule{6-7}
 \cmidrule{8-9}
 \cmidrule{10-11}
CF-Method & A & B & A & B & A & B & A & B & A & B\\
\midrule
Min-Edit  & 0.62 & 0.34 & 0.33 & 0.21 & 1.01 & 1.06 & 0.04 & 0.00 & 42.72 & 41.29 \\
CEM  & 0.59 & 0.19 & 0.33 & 0.13 & 1.00 & 1.10 & 0.04 & 0.00 & 43.17 & 41.46 \\
VLK  & 0.66 & 0.31 & 0.30 & 0.21 & 1.01 & 1.03 & 0.08 & 0.06 & 49.25 & 45.85 \\
Revise  & 0.33 & 0.16 & 0.23 & \textbf{0.03} & 1.04 & \textbf{0.99} & \textbf{0.32} & \textbf{0.12} & 45.76 & 49.42 \\
Human & \textbf{0.94} & \textbf{0.71} & \textbf{0.11} & 0.15 & \textbf{0.98} & 1.02 & 0.06 & 0.05 & \textbf{50.05} & \textbf{55.98}\\
\bottomrule
\end{tabular}}

\end{table}

\subsection{Representativeness: Human Explanations are Within Distribution (the Counterfactual One)}
\label{sec:repeval}

\noindent Counterfactual explanations should be within distribution and, to some degree, representative of the counterfactual class to function as useful explanations. Hence, methods with better within-distribution scores are to be preferred.  But, how do the within-distribution properties of human explanations compare to those of machine explanations?

The Monte Carlo Dropout (MC-Mean, MC-Std)  \cite{gal2016dropout, bhatt2021uncertainty} metric which measures the uncertainty in a model's prediction confidence  \cite{kenny2020generating, schut2021generating, delaney2021uncertainty, bhatt2021uncertainty}, shows that human counterfactuals are the least uncertain with respect to the model's classification \cite{kenny2020generating, gal2016dropout}, whereas all four XAI methods have lower certainty scores. Notably, Revise, which was closest to the human counterfactuals on distance, diverges more than any other method on this measure, indicating that its explanations are distributionally quite different to  the human ground-truth.  In short, humans do not create visual explanations that are close to the model's decision boundary (i.e., ones with high aleatoric uncertainty \cite{schut2021generating}).

Furthermore, the  R\%-sub metric \cite{samangouei2018explaingan} shows that human counterfactuals are prototypical with respect to the counterfactual class; they have the highest  R\%-sub scores showing that they are the most representative of the counterfactual class, relative to class prototypes retrieved using MMD-critic \cite{kim2016criticism}, in contrast to the scores seen for the XAI methods. Finally, the IM1 and LOF metrics confirm this interpretation of where human counterfactuals sit, class-wise. IM1 \cite{VanLooveren2019}, which uses an autoencoder to estimate the reconstruction error for the counterfactual class, shows that human counterfactuals have the lowest error relative to all four XAI methods for MNIST. 10-LOF \cite{breunig2000lof}, which is a proximity based out-of-distribution measure in the pixel space \cite{kanamori2020dace, keane2021if} demonstrates that human explanations are more well grounded in the counterfactual class relative to min-edit counterfactuals.

In short, none of the current XAI methods, whether they be constraint optimisers, autoencoders or generative models  \cite{joshi2019towards, VanLooveren2019, kenny2020generating, Singla2020Explanation, dhurandhar2018explanations}, do a good job of corresponding to the human ground-truth. Human explanations reveal a tendency to produce counterfactuals that can be quite distant from the query, while being close to the prototype(s) of the counterfactual class.

\begin{figure}
\includegraphics[width=1\linewidth]{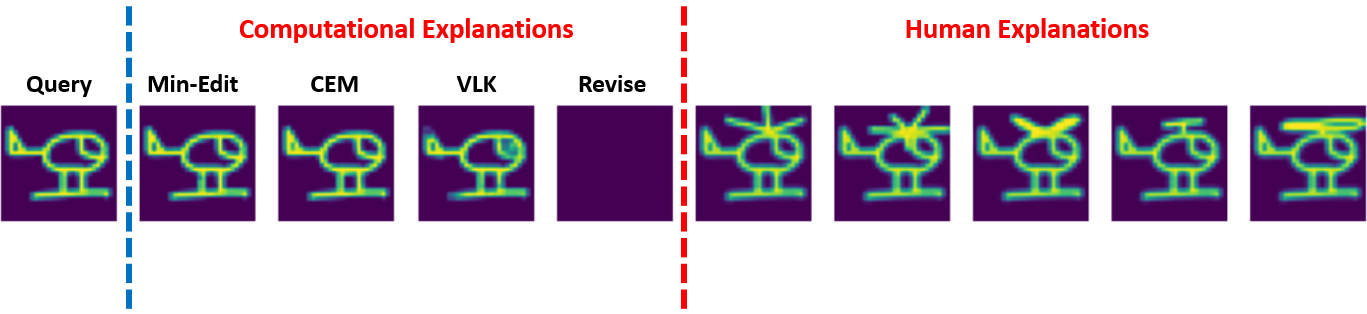}
\caption{\textit{Prototype Evaluations}.  From the QuickDraw data, a query (a ``helicopter" misclassified as a ``mushroom") and the explanations generated by four XAI methods (as four counterfactual ``helicopters") compared to those generated by users.  Note, how people add ``rotor blades", a semantic-feature, whereas the automated methods perform minimal pixel changes. Revise fails to generate an explanation, confirming findings by \cite{holtgen2021deduce}.}

\label{fig:protoeval}
\end{figure}

\subsection{Prototypicality: Human Counterfactuals Are More Prototypical}
\label{sec:proto}

On the face of it, people's counterfactual explanations for misclassified QuickDraw Doodles show semantic-features being added, informed by prototypes in the counterfactual class (i.e., latent features in many CNNs \cite{kim2018interpretability, ghorbani2019towards, chen2020concept, zhang2021invertible}). Figure \ref{fig:protoeval} shows an image of a ``helicopter" that was misclassified as a ``mushroom", to which people add ``rotor blades" to identify it as a ``helicopter"\footnote{Interestingly, in both user studies people could add pixels or erase them when producing their explanations, but they tended to add pixels more often than remove them, echoing known psychological findings. Byrne \cite{Byrne2019} noted that \textit{``people tend to create counterfactuals about how things could have been different, that add something new to what they already know about the situation, rather than ones that remove something from it''}}.  In contrast, the XAI methods make small changes to a few pixels that imperceptibly modify the image.

This role of prototypes in counterfactual explanations can be evaluated more directly by analysing similarities in the latent space.  We used the Grad-Cos similarity metric \cite{charpiat2019input, hanawa2021evaluation} to compare human counterfactuals to class prototypes from the counterfactual class (retrieved using MMD-Critic \cite{kim2016criticism}).  As Figure 6 shows, in this latent space, human counterfactuals are more similar than all four XAI methods, to the prototypes of the counterfactual class. These results confirm the intuition that people modify the semantic-features of images in producing counterfactual explanations, shaping these explanations relative to the prototypes of the counterfactual class.

\begin{figure}
\centering
\begin{subfigure}{.5\textwidth}
  \centering
  \includegraphics[width=1\linewidth]{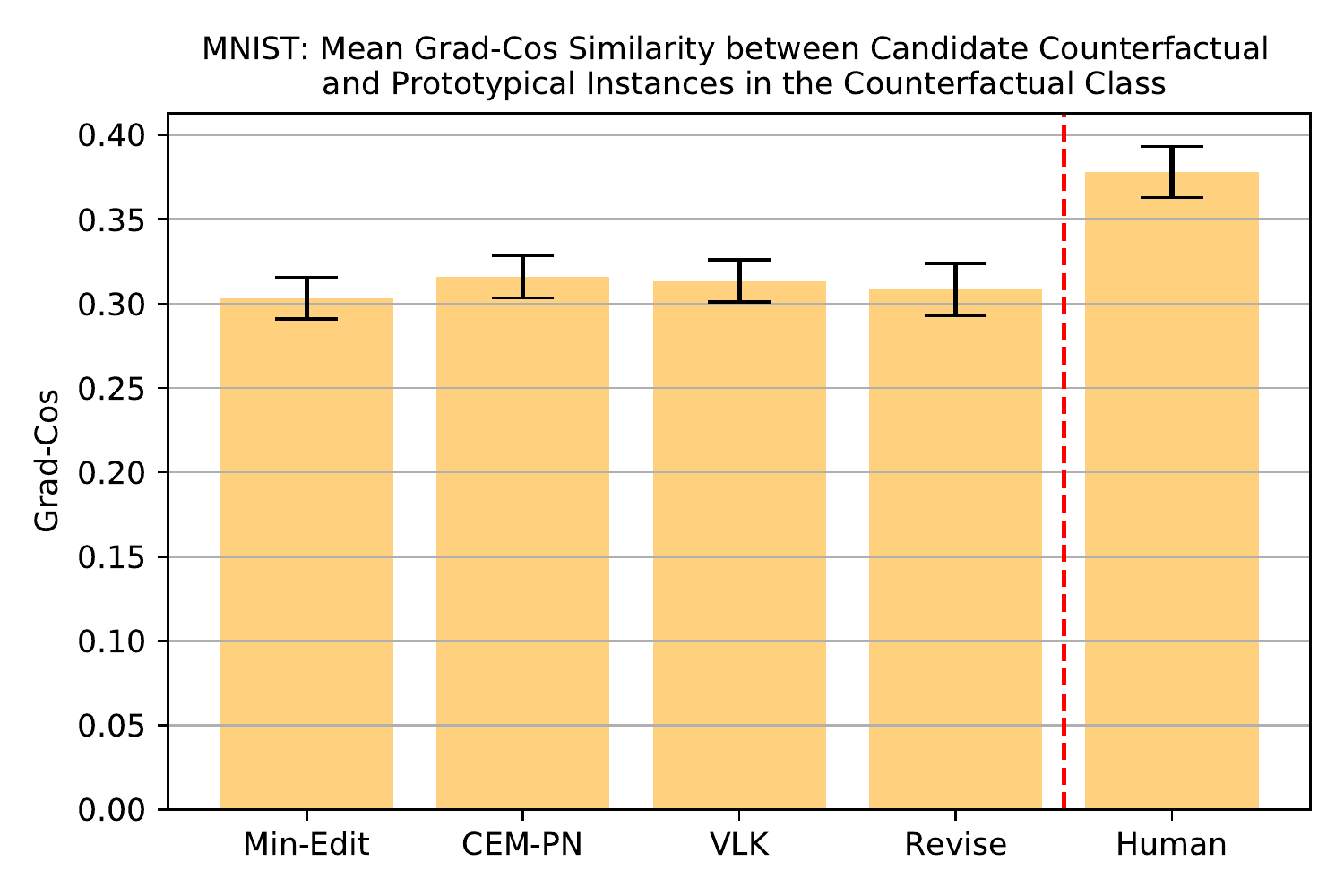}
  \caption{MNIST}
  \label{fig:sub1}
\end{subfigure}%
\begin{subfigure}{.5\textwidth}
  \centering
  \includegraphics[width=1\linewidth]{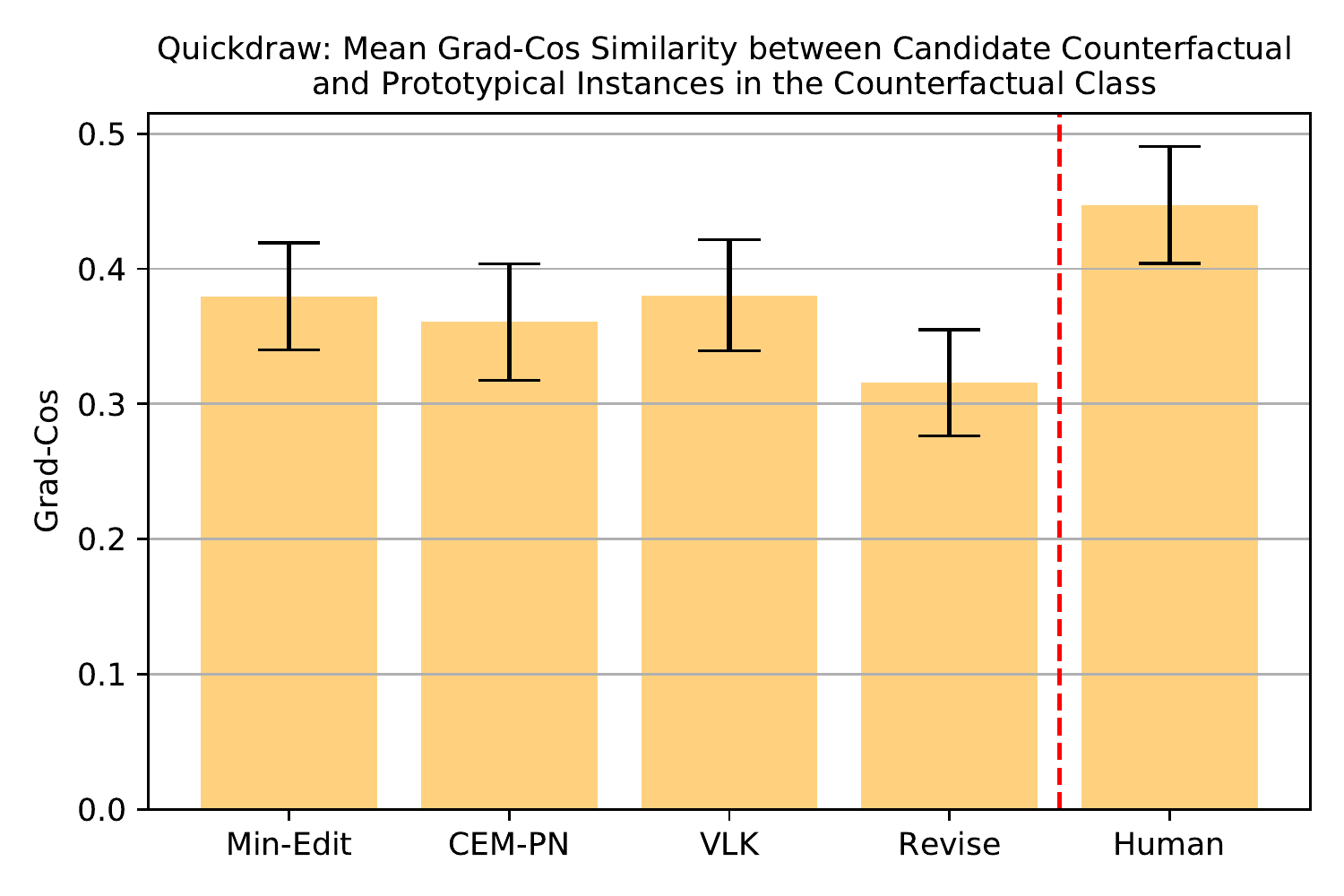}
  \caption{QuickDraw}
  \label{fig:sub2}
\end{subfigure}
\caption{\textit{Prototype Evaluations}. Mean Grad-Cos Similarity scores for counterfactual-prototype and query-prototype pairs (prototypes retrieved using MMD-critic) from XAI methods compared to the human ground-truth, for (a) MNIST and (b) QuickDraw datasets (Error bars show standard error of the mean).} 

\label{fig:test}
\end{figure}

\begin{figure}
\includegraphics[width=1\linewidth]{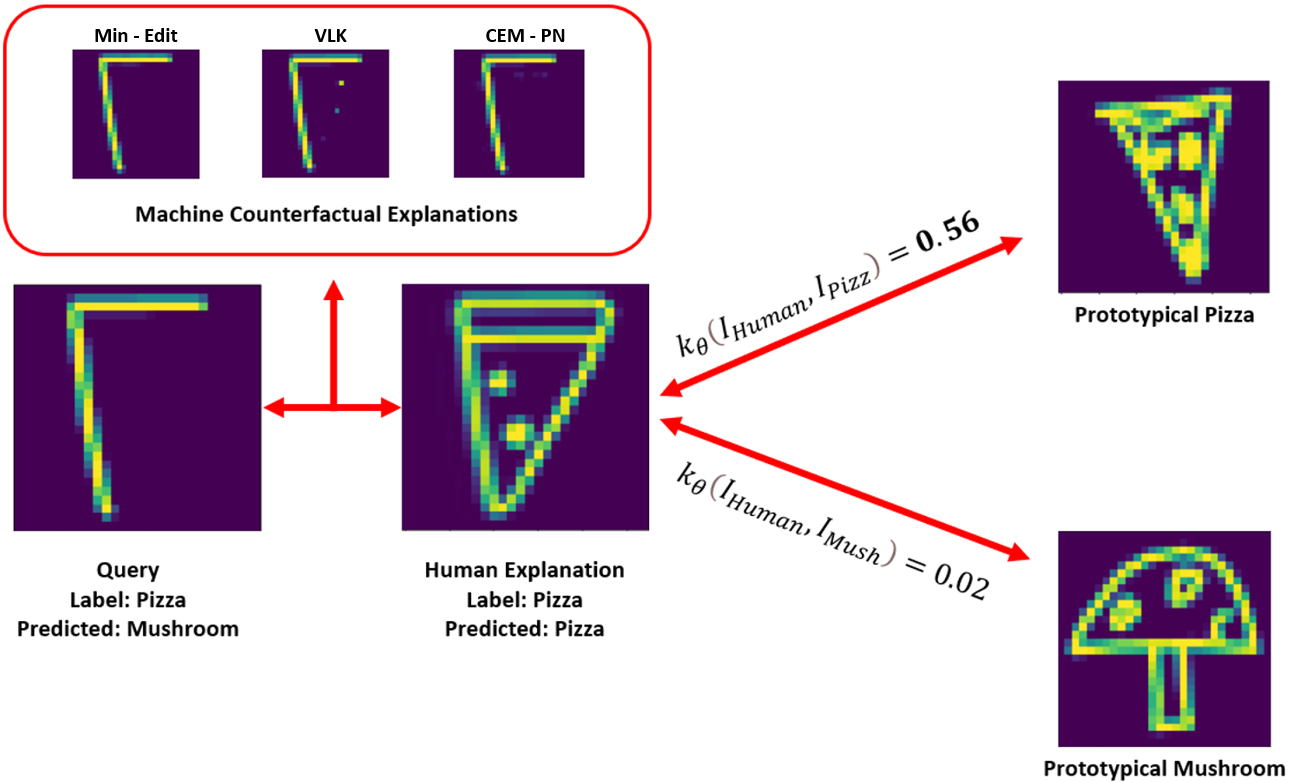}
\caption{As the original QuickDraw images were created by humans under a 20 second time constraint, some  are unfinished, presenting interpretation and reconstruction difficulties for many XAI models (see insert).  In contrast, people readily fill in missing portions of the image, presumably using their knowledge of the class prototype (e.g., see similarities to MMD-Critic prototypes)}
\label{fig:quickdraw}
\end{figure}

Humans generate counterfactual explanations that are more similar to class prototypes in the latent space, relative to other computational methods for both MNIST and QuickDraw data-sets. In the case of the QuickDraw dataset, the original QuickDraw image doodles were produced under a strict time constraint of 20 seconds, which meant that some doodles were left unfinished. These unfinished, noisy doodles present significant challenges for the XAI methods, as they are degraded inputs (and possibly corrupt the ability of the Revise and VLK generative models to reconstruct the class distribution). However, people manage these degraded inputs using their knowledge of the semantic features of the represented objects. Figure \ref{fig:quickdraw} shows the case of an unfinished doodle of a pizza, which the model predicts to be a mushroom. All the XAI methods fail to generate plausible counterfactuals to this image, yet people simply add significant missing semantic features, leveraging their knowledge of class prototypes.

\section{Conclusion}
\label{sec:summary}

Although many different methods for counterfactual explanation have been developed in the XAI literature (e.g., \cite{karimi2020algorithmiccf,keane2021if}), the paucity of user studies in this field has left most claims of psychological validity untested. The present work addresses this deficit by gathering human ground-truths on counterfactual explanation for two benchmark image datasets (i.e., MNIST, QuickDraw Doodles).  It then used benchmark metrics (e.g., on distance, representativeness and latent similarity) to compare machine-generated explanations to those that people actually use when explaining counterfactually.  Hence, for the first time, the extent of the divergence between XAI methods and human, ground-truth performance can be accurately assessed.  The results show that, unlike machine methods, human explanations focus more on semantic features and are closer to prototypes in the counterfactual class, than those of the query class. As such, this work establishes key, novel human benchmarks for assessing the plausibility of XAI methods. Lewis' \cite{lewis2013counterfactuals} philosophical analysis of counterfactuals argued that they were the closest possible world to the current one based on minimal differences. The present work shows that people compute those minimal differences in a semantic space, rather than a pixel space, and do so with an eye to the counterfactual world, rather than the current one.

\section{\textbf{Data and Code Availability}}
The MNIST dataset is openly available\footnote{\url{http://yann.lecun.com/exdb/mnist/}} and the Google QuickDraw dataset is made available by Google, Inc. under the Creative Commons Attribution 4.0 International license\footnote{\url{https://github.com/googlecreativelab/quickdraw-dataset}}. All data from the user study is made available \footnote{\url{https://github.com/e-delaney/cfe_images_how_people_differ_from_machines/tree/main/User_data}} and the code used to detail hyperparamaters and produce our results is also provided\footnote{\url{https://github.com/e-delaney/cfe_images_how_people_differ_from_machines}}.

\section{\textbf{Acknowledgements}}
This publication has emanated from research conducted
with the financial support of (i) Science Foundation Ireland
(SFI) to the Insight Centre for Data Analytics under Grant
Number 12/RC/2289 P2 and (ii) SFI and the Department of
Agriculture, Food and Marine on behalf of the Government
of Ireland under Grant Number 16/RC/3835 (VistaMilk).
Thanks also to Greta Warren for the Power analysis in the literature review of Goyal, and the ideas on subjective and objective measures.

\clearpage
\appendix
\section{\textbf{Monte Carlo Dropout Technical Details}}
\label{Appendix_MC}
Following the description of \cite{AWS_Uncertainty, delaney2021uncertainty}, we provide an overview of how MC-Dropout can be applied. When a predictive distribution $p(y \vert x, D)$ is obtained, the corresponding uncertainty can be uncovered by analysing the variance.

From a Bayesian perspective, total predictive uncertainty, ${\mathbb{V}(y \mid x)}$, can be decomposed into a sum of two components, namely epistemic (model) uncertainty and aleatoric (data) uncertainty \cite{kendall2017uncertainties, AWS_Uncertainty}. Let $x$ represent some input, $y$ the target variable and $\Theta$ the random parameters of the model, then
\begin{equation}
        {\mathbb{V}(y \mid x)} = \underbrace{\mathbb{V}(\mathbb{E}(y \mid x,\Theta))}_{Epistemic} + \underbrace{{\mathbb{E}(\mathbb{V}(y \mid x,\Theta))}}_{Aleatoric}
\end{equation}
To learn this distribution we learn the distribution over the model parameters p($\Theta \vert D$) (i.e., the parametric posterior distribution).

Work in \cite{gal2016dropout} showed that, by randomly switching off neurons in a neural network using different dropout configurations, one could approximate the parametric posterior distribution without the need to retrain the network. Each dropout configuration $\Theta_{t}$ corresponds to a sample from the approximate parametric posterior distribution $q(\Theta \vert D)$ s.t. $\Theta_{t} \sim q(\Theta \vert D)$.

Sampling from the approximate posterior enables us to uncover the predictive distribution $p(y \mid x)$:
\begin{equation}
     p(y \mid x, D) \approx   \int_\Omega \underbrace{p(y \mid x, \Theta)}_{likelihood} \underbrace{q(\Theta \mid D}_{posterior}) \,d\Theta
\end{equation}
The above integral can be approximated through Monte Carlo methods, giving;
\begin{equation}
     p(y \mid x ,D) \underbrace{\approx}_{MC}   \frac{1}{T}\sum_{t=1}^{T} p(y \mid x, \Theta_{t})
\end{equation}
Multiple forward passes with different dropout configurations allow one to uncover the predictive distribution. Under the assumption that the likelihood is Gaussian distributed, the mean $f(x, \theta)$ and variance $s^{2}(x, \Theta)$ parameters of the Gaussian function are determined from Monte Carlo simulation such that $f(x, \theta), s^{2}(x, \Theta) \sim$ MC-Dropout(x), and can yield useful information about the predictive uncertainty through connecting back with Equation A.2.
\label{mcdropout_method}

\section{Prototype Evaluations: Prototypes \& Similarity}
\label{sec:protosim}

To determine the closeness of generated counterfactuals to the prototype(s) of the counterfactual class, MMD-Critic \cite{kim2016criticism} was used to generate prototypes for the class and then Grad-Cos was used to measure the latent similarity between explanations and prototypes in order to determine if explanations generated by humans are more similar to class prototypes relative to explanations that are automatically generated. MMD-critic is briefly described below.

\paragraph{\textbf{Prototype Retrieval: MMD-Critic}}
Introduced by Kim et al. \cite{kim2016criticism}, this approach computes prototypes by minimizing the maximum mean discrepancy between the prototype distribution and the data distribution. These densities are estimated using a kernel density function, $k$. Following \cite{molnar2020, kim2016criticism}, let $m$ represent the number of individual prototypes $z$ and $n$ represent the number of data-points $x$ in the dataset. Then the $MMD^{2}$ can be represented by:
\begin{equation}
MMD^2=\frac{1}{m^2}\sum_{i,j=1}^m{}k(z_i,z_j)-\frac{2}{mn}\sum_{i,j=1}^{m,n}k(z_i,x_j)+\frac{1}{n^2}\sum_{i,j=1}^n{}k(x_i,x_j)
\end{equation}
The first term calculates the average proximity of the prototypes to each other, while the last term calculates the average proximity of the data-points to each other. The middle term calculates the average proximity between the prototypes and the other data-points (multiplied by 2). In our implementation we use a standard radial basis function as our choice for the kernel $k$, defined by:
\begin{equation}
k(x,x')=exp\left(-\gamma\||x-x'\||_{2}\right)
\end{equation}

\noindent The $MMD^{2}$ measure, kernel function and greedy search are combined in an algorithm to find prototypes \cite{molnar2020}. Starting with an empty list of prototypes, each point in the class are evaluated using $MMD^{2}$, and the point that minimizes $MMD^{2}$ to the largest degree is added to the list.

\section{Additional User Study Details and Results}
\label{Appendix_user}

\begin{figure}[h!]
 \centering
 \begin{subfigure}{.5\textwidth}
   \centering
   \includegraphics[width=1\linewidth]{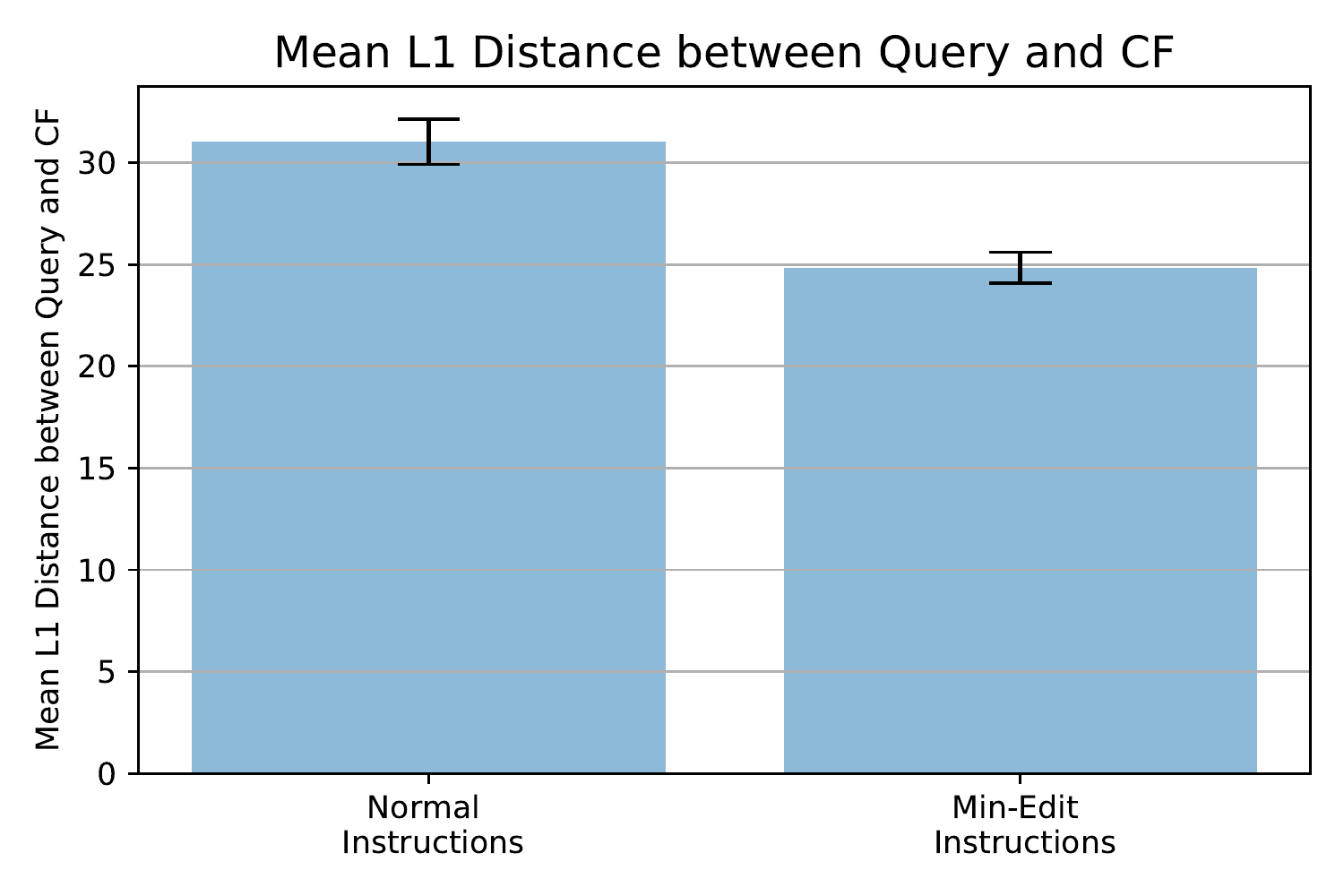}
   \label{fig:distances1}
 \end{subfigure}%
 \begin{subfigure}{.5\textwidth}
   \centering
   \includegraphics[width=1\linewidth]{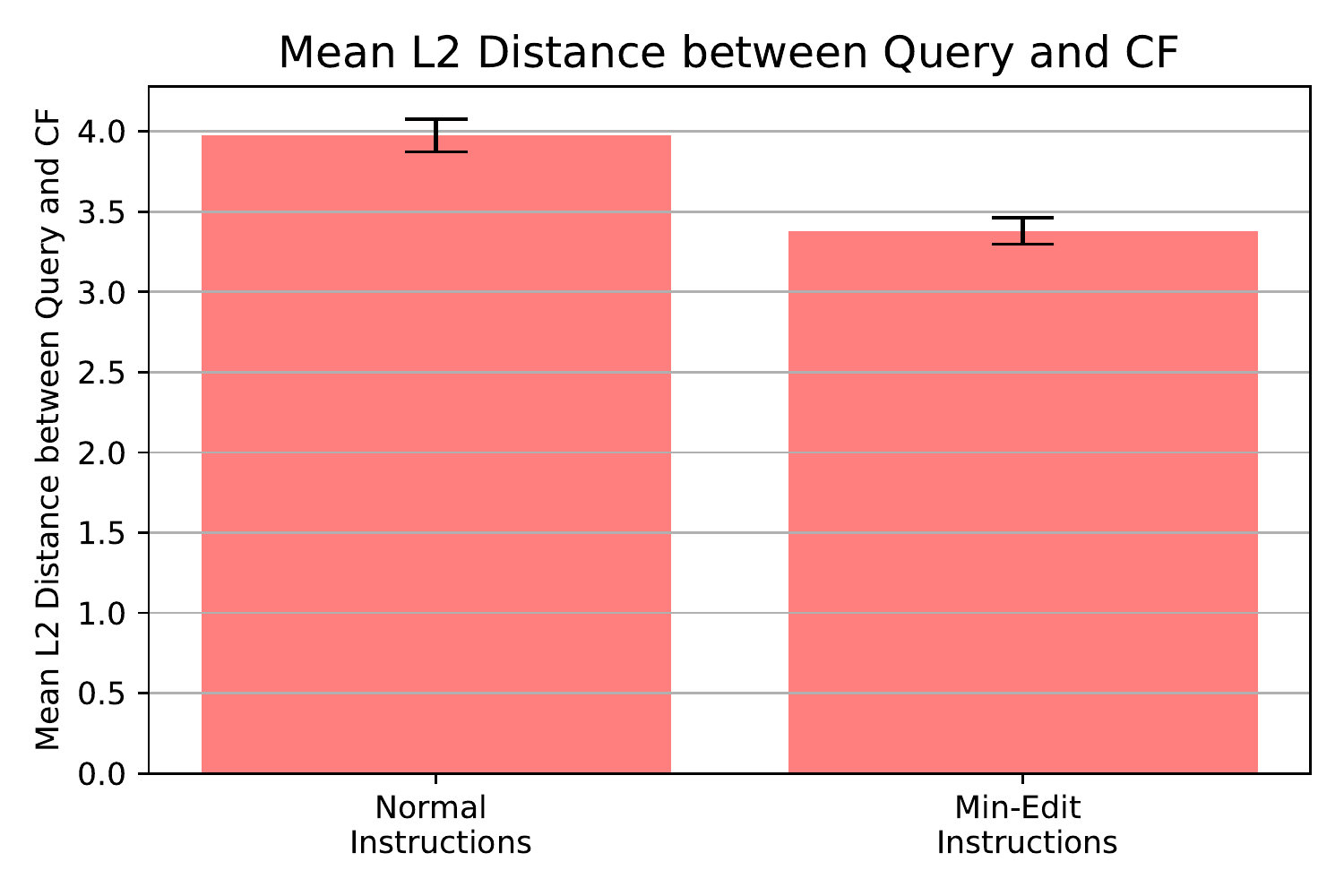}
   \label{fig:distances2}
 \end{subfigure}
 \caption{\textit{Proximity Evaluations}. Mean L1 and L2 distances between the misclassified query and the corresponding counterfactual. Explanations produced by the two human groups -- Normal (people who weren't given any specific instructions on how to edit the image) \& Min-Edit (people who were specifically instructed to make the smallest edit possible) -- for MNIST (error bars show standard error of the mean).}
 \label{fig:distances_appendix}
 \end{figure}

Unsurprisingly, users who were instructed to minimally edit the query to generate a counterfactual explanation, made significantly smaller edits (according to L1 and L2 distance between the query and generated counterfactual, two-sample t-test, $p < 0.001$) relative to users who were not given any specific instructions on how to modify the image (See Figure \ref{fig:distances_appendix}).

\clearpage

\bibliographystyle{elsarticle-num-names-alphsort.bst}
\bibliography{elsarticle-template-num-names}





\end{document}